\documentclass[5p,times,twocolumn]{elsarticle}

\usepackage{hyperref}

\usepackage{picins}
\usepackage{color}
\usepackage{booktabs} 
\usepackage{amsmath,amssymb}
\usepackage{algorithm}
\usepackage{algorithmic}
\usepackage{graphicx} 
\usepackage{subfigure} 
\usepackage{verbatim}
\usepackage{psfrag}
\usepackage{xspace}
\usepackage{balance}

\newcommand{\vct}[1]{\ensuremath{\boldsymbol{#1}}} 

\newcommand{\set}[1]{\ensuremath{\mathcal{#1}}}

\newcommand{\argmax}{\operatornamewithlimits{\arg\,\max}}

\newcommand{\argmin}{\operatornamewithlimits{\arg\,\min}}

\newcommand{\myparagraph}[1]{\noindent \textbf{#1.}}

\newcommand{\ie}{{i.e.}\xspace}
\newcommand{\eg}{{e.g.}\xspace}
\newcommand{\etal}{{et al.}\xspace}
\newcommand{\etc}{{etc.}\xspace}

\journal{Pattern Recognition}

\biboptions{sort&compress}

\begin{document}

\begin{frontmatter}

\title{Wild Patterns: Ten Years After the Rise of \\Adversarial Machine Learning}

\author[myprimaryaddress,mysecondaryaddress]{Battista Biggio\corref{c}}
\cortext[c]{Corresponding author}
\ead{battista.biggio@diee.unica.it}

\author[myprimaryaddress,mysecondaryaddress]{Fabio Roli}
\ead{roli@diee.unica.it}

\address[myprimaryaddress]{Department of Electrical and Electronic Engineering, University of Cagliari, Italy}



\address[mysecondaryaddress]{Pluribus One, Cagliari, Italy}


\begin{abstract}
Learning-based pattern classifiers, including deep networks, have shown impressive performance in several application domains, ranging from computer vision to cybersecurity. 
However, it has also been shown that adversarial input perturbations carefully crafted either at training or at test time can easily subvert their predictions.
The vulnerability of machine learning to such wild patterns (also referred to as adversarial examples), along with the design of suitable countermeasures, have been investigated in the research field of adversarial machine learning.
In this work, we provide a thorough overview of the evolution of this research area over the last ten years and beyond, starting from pioneering, earlier work on the security of non-deep learning algorithms up to more recent work aimed to understand the security properties of deep learning algorithms, in the context of computer vision and cybersecurity tasks.
We report interesting connections between these apparently-different lines of work, highlighting common misconceptions related to the security evaluation of machine-learning algorithms.
We review the main threat models and attacks defined to this end, and discuss the main limitations of current work,  along with the corresponding future challenges towards the design of more secure learning algorithms.
\end{abstract}


\end{frontmatter}


\section{Introduction} \label{sect:intro}
\vspace{-5pt}
Modern technologies based on pattern recognition, machine learning and data-driven artificial intelligence, especially after the advent of deep learning, have reported impressive performance in a variety of application domains, from classical pattern recognition tasks like speech and object recognition, used by self-driving cars and robots, to more modern cybersecurity tasks like spam and malware detection~\cite{gu18-pr}.\footnote{The term \emph{malware} (short for \emph{mal}icious softw\emph{are}) is normally used to refer to harmful computer programs in general, including computer viruses, worms, ransomware, spyware, \etc}
It has been thus surprising to see that such technologies can easily be fooled by \emph{adversarial examples}, \ie, carefully-perturbed input samples aimed to mislead detection at test time.
This has brought considerable attention since 2014, when Szegedy~\etal~\cite{szegedy14-iclr} and subsequent work~\cite{goodfellow15-iclr,nguyen15-cvpr,moosavi16-deepfool} showed that 
deep networks for object recognition can be fooled by input images perturbed in an imperceptible manner. 

Since then, an ever-increasing number of research papers have started proposing countermeasures to mitigate the threat associated to these \emph{wild patterns}, not only in the area of computer vision~\cite{papernot16-distill,melis17-vipar,meng17-ccs,lu17-iccv,li17-iccv,grosse17-esorics,jordaney17-usenix}.\footnote{More than 150 papers on this subject were published on ArXiv only in the last two years.}
This huge and growing body of work has clearly fueled a renewed interest in the research field known as \emph{adversarial machine learning}, while also raising a number of misconceptions on how the security properties of learning algorithms should be evaluated and understood.

The primary misconception is about the start date of the field of \emph{adversarial machine learning}, which is not 2014. This wrong start date is implicitly acknowledged in a growing number of recent papers in the area of computer security~\cite{mcdaniel16,papernot16-sp,papernot16-distill,papernot17-asiaccs,grosse17-esorics,carlini17-sp,meng17-ccs,sharif16-ccs} and computer vision~\cite{lu17-iccv,li17-iccv,xie17-iccv}, which focus mainly on the study of security and robustness of deep networks to adversarial inputs.
However, as we will discuss throughout this manuscript, this research area has been independently developing and re-discovering well-known phenomena that had been largely explored in the field of adversarial machine learning before the discovery of adversarial examples against deep networks.

To the best of our knowledge, the very first, seminal work in the area of adversarial machine learning dates back to 2004.
At that time, Dalvi~\etal~\cite{dalvi04}, and immediately later Lowd and Meek~\cite{lowd05,lowd05-ceas} studied the problem in the context of spam filtering, showing that linear classifiers could be easily tricked by few carefully-crafted changes in the content of spam emails, without significantly affecting the readability of the spam message. These were indeed the first adversarial examples against linear classifiers for spam filtering. Even earlier, Matsumoto~\etal~\cite{matsumoto02} showed that fake fingerprints can be fabricated with plastic-like materials to mislead biometric identity recognition systems.
In 2006, in their famous paper, Barreno~\etal~\cite{barreno06-asiaccs} questioned the suitability of machine learning in adversarial settings from a broader perspective, categorizing attacks against machine-learning algorithms both at training and at test time, and envisioning potential countermeasures to mitigate such threats. Since then, and independently from the discovery of \emph{adversarial examples} against deep networks~\cite{szegedy14-iclr}, a large amount of work has been done to: 
($i$) develop attacks against machine learning, both at training time (poisoning)~\cite{nelson08,rubinstein09,biggio12-icml,kloft10,kloft12b,biggio15-icml,biggio14-svm-chapter,mei15-aaai,koh17-icml,biggio17-aisec} and at test time (evasion)~\cite{wittel04,dalvi04,lowd05,lowd05-ceas,globerson06-icml,teo08,dekel10,biggio13-ecml,srndic14,biggio14-svm-chapter};
($ii$) propose systematic methodologies for security evaluation of learning algorithms against such attacks~\cite{barreno10,biggio14-tkde,biggio14-ijprai,biggio15-spmag,srndic14};
and ($iii$) design suitable defense mechanisms to mitigate these threats~\cite{dalvi04,globerson06-icml,kolcz09,biggio08-spr,bruckner12,bulo17-tnnls,demontis17-tdsc,jordaney17-usenix}.

The fact that \emph{adversarial machine learning} was well-established before 2014 is also witnessed by a number of related events, including the 
2007 NIPS Workshop on Machine Learning in Adversarial Environments for Computer Security~\cite{nips07-adv}, along with the subsequent special issue on the journal \emph{Machine Learning}~\cite{laskov10-ed}, the 2013 Dagstuhl Perspectives Workshop on Machine Learning Methods for Computer Security~\cite{joseph13-dagstuhl} and, most importantly, the Workshop on Artificial Intelligence and Security (AISec), which reached its 10th edition in 2017~\cite{biggio17-aisec-proc}. Worth remarking, a book has also been recently published on this subject~\cite{joseph18-advml-book}.

In this work, we aim to provide a thorough overview of the evolution of this interdisciplinary research area over the last ten years and beyond, from pioneering work on the security of (non-deep) learning algorithms to more recent work focused on the security properties of deep learning algorithms, in the context of computer vision and cybersecurity tasks.
Our goal is to \emph{connect the dots} between these apparently-different lines of work, while also highlighting common misconceptions related to the security evaluation of learning algorithms.

We first review the notion of arms race in computer security, advocating for a proactive security-by-design cycle that explicitly accounts for the presence of the attacker in the loop (Sect.~\ref{sect:arms-race}).
Our narrative of the security of machine learning then follows three metaphors, referred to as the \emph{three golden rules} in the following: ($i$) know your adversary, ($ii$) be proactive; and ($iii$) protect yourself.
Knowing the attacker amounts to modeling threats against the learning-based system under design. To this end, we review a comprehensive threat model which allows one to envision and simulate attacks against the system under design, to thoroughly assess its security properties under well-defined attack scenarios (Sect.~\ref{sect:attack-framework}).
We then discuss how to proactively simulate test-time evasion and training-time poisoning attacks against the system under design (Sect.~\ref{sect:attacks}), and how to protect it with different defense mechanisms (Sect.~\ref{sect:defenses}).
We finally discuss the main limitations of current work and the future research challenges towards the design of more secure learning algorithms (Sect.~\ref{sect:conclusions}).

\vspace{-5pt}
\section{Arms Race and Security by Design}
\label{sect:arms-race}
\vspace{-5pt}
Security is an arms race, and the security of machine learning and pattern recognition systems is not an exception to this~\cite{biggio14-tkde,biggio14-ijprai,corona13-is}.
To better understand this phenomenon, consider that, since the 90s, computer viruses and Internet scams have increased not only in terms of absolute numbers, but also in terms of variability and sophistication, in response to the growing complexity of defense systems. 
Automatic tools for designing novel variants of attacks have been developed, making large-scale  automatization of stealthier attacks practical also for non-skilled attackers. A very clear example of this is provided by \emph{phishing kits}, which automatically compromise legitimate (vulnerable) websites in the wild, and hide phishing webpages within them~\cite{han16-ccs,corona17-esorics}.
The sophistication and proliferation of such attack vectors, malware and other threats, is strongly motivated by a flourishing underground economy, which enables easy monetization after attack.
To tackle the increasing complexity of modern attacks, and favor the detection of never-before-seen ones, machine learning and pattern recognition techniques have been widely adopted over the last decade also in a variety of cybersecurity application domains.
However, as we will see throughout this paper, machine learning and pattern recognition techniques turned out not to be the definitive answer to such threats. They introduce specific vulnerabilities that skilled attackers can exploit to compromise the whole system, \ie, machine learning itself can be the \emph{weakest link} in the security chain.

To further clarify how the aforementioned arms race typically evolves, along with the notions of reactive and proactive security, we briefly summarize in the following an exemplary case in spam filtering.

\begin{figure*}[t]
\begin{center}
\includegraphics[width=0.4\textwidth]{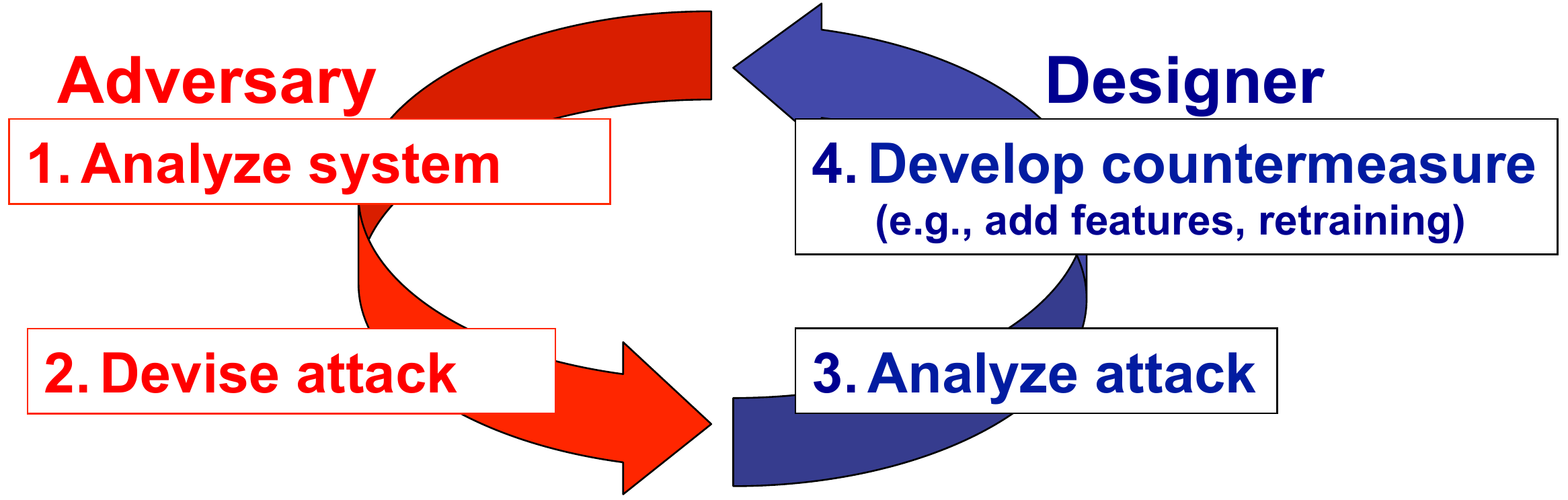}
\hspace{5pt}
\includegraphics[width=0.4\textwidth]{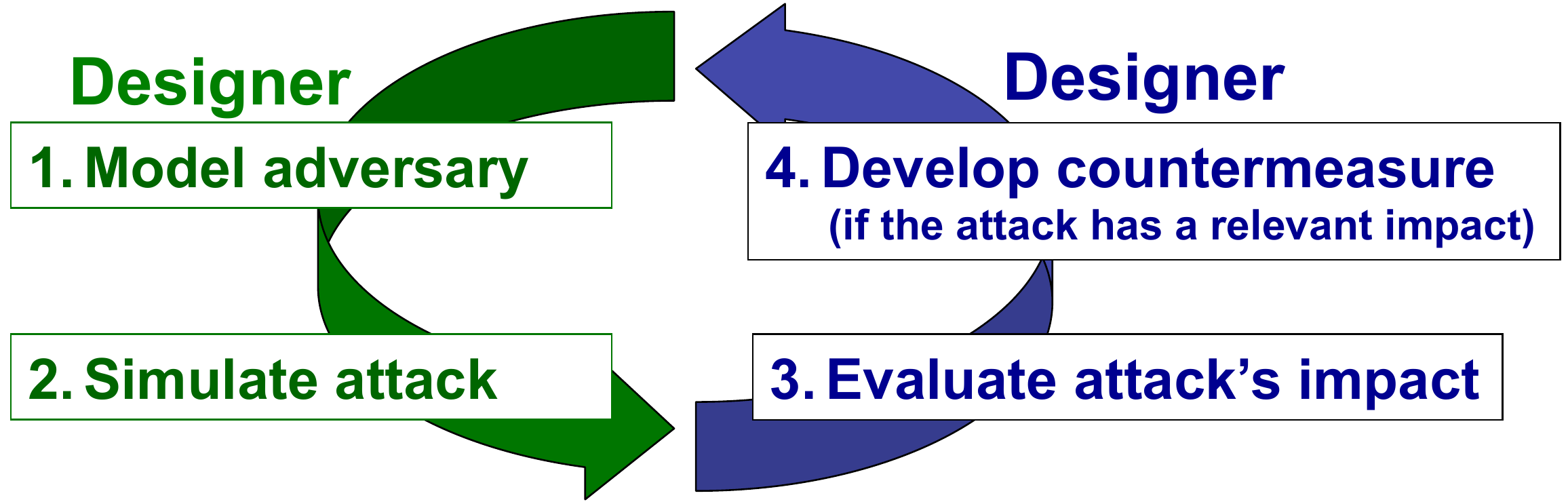}
\caption{A conceptual representation of the reactive (left) and proactive (right) arms races for pattern recognition and machine learning systems in computer security~\cite{biggio14-tkde,biggio14-ijprai}.}
\label{fig:arms-race}
\end{center}
\end{figure*}

\begin{figure}[t]
\centering
\fboxsep=0mm\fbox{\includegraphics[width=0.33\textwidth, height=0.15\textwidth]{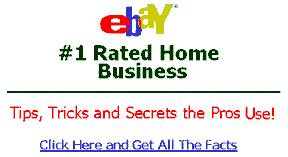}}\\
\hspace{2pt}\fboxsep=0mm\fbox{\includegraphics[width=0.33\textwidth, height=0.15\textwidth]{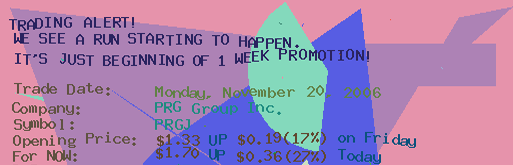}}
\caption{Examples of \emph{clean} (top) and \emph{obfuscated} (bottom) spam images~\cite{biggio11-prl}.}
\vspace{-8pt}
\label{fig:image-spam}
\end{figure}

\myparagraph{The spam arms race} Spam emails typically convey the spam message in textual format, which can be detected by rule-based filters and text classifiers. Spammers  attempt to mislead these defenses by obfuscating the content of spam emails to evade detection, \eg, by misspelling \emph{bad} words (\ie, words likely to appear in spam but not in legitimate emails), and adding \emph{good} words (\ie, words typically occurring in legitimate emails, randomly guessed from a reference vocabulary)~\cite{wittel04,kolcz09,lowd05-ceas}.
In 2005, spammers invented a new trick to evade textual-based analysis, referred to as \emph{image-based spam} (or image spam, for short)~\cite{biggio11-prl,attar13}. The idea is simply to embed the spam message within an attached image (Fig.~\ref{fig:image-spam}, left). 
Due to the large amount of image spam sent in 2006 and 2007, countermeasures were promptly developed based on signatures of known spam images (through hashing), and on extracting text from suspect images with OCR tools~\cite{fumera06}.
To evade these defenses, spammers started obfuscating images with random noise patterns (Fig.~\ref{fig:image-spam}, right) that, ironically, were similar to those used in CAPTCHAs to protect web sites from spam bots~\cite{thomas09}.
Learning-based approaches based on low-level visual features were then devised to detect spam images. Image spam volumes have since declined, but spammers have been constantly developing novel tricks to evade detection.

\myparagraph{Reactive and proactive security} As discussed for spam filtering,
security problems are often cast as a \emph{reactive} arms race, in which the system designer and the attacker aim to achieve their goals by adapting their behavior in response to that of the opponent, \ie, \emph{learning from the past}.
This can be modeled according to the following
steps (Fig.~\ref{fig:arms-race}, left)~\cite{biggio14-tkde,biggio14-ijprai}: ($i$) the attacker analyzes the
defense system and crafts an attack to violate its security; and ($ii$) the system designer analyzes the newly-deployed attacks and designs novel countermeasures against them.
However, \emph{reactive} approaches are clearly not able to prevent the risk of \emph{never-before-seen} attacks.
To this end, the designer should follow a \emph{proactive} approach to anticipate the attacker by ($i$) identifying relevant threats against the system under design and simulating the corresponding attacks, ($ii$) devising suitable countermeasures (if retained necessary), and ($iii$) repeating this process \emph{before} system deployment (Fig.~\ref{fig:arms-race}, right).
In practice, these steps are facilitated by leveraging a thorough model of the attacker, as that discussed in the next section, which helps envisioning and analyzing a number of potential attack scenarios against learning-based systems.

\section{Know Your Adversary: Modeling Threats}
\label{sect:attack-framework}
\vspace{-5pt}
\begin{quote}
\begin{flushright}
\emph{``If you know the enemy and know yourself, you need not fear the result of a hundred battles.''}
(Sun Tzu, The Art of War, 500 BC)
\end{flushright}
\end{quote}

We discuss here the \emph{first golden rule} of the proactive security cycle discussed in the previous section, \ie, how to model threats against learning-based systems and thoroughly evaluate their security against the corresponding attacks.
To this end, we exploit a framework based on the popular attack taxonomy proposed in~\cite{barreno06-asiaccs,barreno10,huang11} and subsequently extended in~\cite{biggio14-tkde,biggio14-ijprai,biggio14-svm-chapter,melis17-vipar,biggio17-aisec}, which enables one to envision different attack scenarios against learning algorithms and deep networks, and to implement the corresponding attack strategies.
Notably, these attacks include training-time poisoning and test-time evasion attacks (also recently referred to as adversarial training and test examples)~\cite{huang11,biggio14-tkde,biggio13-ecml,biggio12-icml,biggio15-icml,mei15-aaai,biggio17-aisec,goodfellow15-iclr,nguyen15-cvpr,moosavi16-deepfool,papernot16-sp,papernot17-asiaccs,carlini17-sp,koh17-icml}.
It consists of defining the attacker's goal, knowledge of the targeted system, and capability of manipulating the input data, to subsequently define an optimization problem corresponding to the optimal attack strategy. The solution to this problem provides a way to manipulate input data to achieve the attacker's goal.
While this framework only considers attacks against \emph{supervised} learning algorithms, we refer the reader to similar threat models to evaluate the security of clustering~\cite{biggio13-aisec,biggio14-aisec,biggio14-spr}, and feature selection algorithms~\cite{biggio15-icml,zhang16-tcyb} under different attack settings.

\myparagraph{Notation} In the following, we denote the sample and label spaces with $\set X$ and $\set Y$, respectively, and the training data with $\set D = (\vct x_{i}, y_{i} )_{i=1}^{n}$, being $n$ the number of training samples. 
We use $L(\set D, \vct w)$ to denote the \emph{loss} incurred by the classifier $f : \set X \mapsto \set Y$ (parameterized by $\vct w$) on $\set D$.
We assume that the classification function $f$ is learned by minimizing an objective function $\set L(\set D, \vct w)$ on the training data. Typically, this is an estimate of the generalization error, obtained by the sum of the empirical loss $L$ on $\set D$ and a regularization term.

\subsection{Attacker's Goal} \label{sect:goal}
This aspect is defined in terms of the desired security violation, attack specificity, and error specificity, as detailed below.

\myparagraph{Security Violation} The attacker may aim to cause: an \emph{integrity} violation, to evade detection without compromising normal system operation; an \emph{availability} violation, to compromise the normal system functionalities available to legitimate users; or a \emph{privacy} violation, to obtain private information about the system, its users or data by reverse-engineering the learning algorithm.

\myparagraph{Attack Specificity} It ranges from \emph{targeted} to \emph{indiscriminate}, respectively, depending on whether the attacker aims to cause misclassification of a specific set of samples (to target a \emph{given} system user or protected service), or of any sample (to target \emph{any} system user or protected service).

\myparagraph{Error Specificity} It can be \emph{specific}, if the attacker aims to have a sample misclassified as a specific class; or \emph{generic}, if the attacker aims to have a sample misclassified as any of the classes different from the true class.\footnote{In~\cite{papernot16-sp}, the authors defined \emph{targeted} and \emph{indiscriminate} attacks (at test time) depending on whether the attacker aims to cause \emph{specific} or \emph{generic} errors. Here we do not follow their naming convention, as it can cause confusion with the interpretation of \emph{targeted} and \emph{indiscriminate} attack specificity also introduced in previous work~\cite{barreno06-asiaccs,barreno10,huang11,biggio14-tkde,biggio14-ijprai,biggio14-svm-chapter,biggio15-icml,biggio13-aisec,biggio14-spr,biggio14-aisec}.}

\subsection{Attacker's Knowledge} \label{sect:knowledge}

The attacker can have different levels of knowledge of the targeted system, including: ($k.i$) the training data $\set D$; ($k.ii$) the feature set $\set X$; ($k.iii$) the learning algorithm $f$, along with the objective function $\set L$ minimized during training; and, possibly, ($k.iv$) its (trained) parameters/hyper-parameters $\vct w$.
The attacker's knowledge can thus be characterized in terms of a space $\Theta$, whose elements encode the components ($k.i$)-($k.iv$) as $\vct \theta=(\set D, \set X, f, \vct w)$.
Depending on the assumptions made on ($k.i$)-($k.iv$), one can describe different attack scenarios.

\myparagraph{Perfect-Knowledge (PK) White-Box Attacks} Here the attacker is assumed to know everything about the targeted system, \ie, $\vct \theta_{\rm PK}=(\set D, \set X, f, \vct w)$. This setting allows one to perform a worst-case evaluation of the security of learning algorithms, providing empirical upper bounds on the performance degradation that may be incurred by the system under attack.

\myparagraph{Limited-Knowledge (LK) Gray-Box Attacks} 
One may consider here different settings, depending on the attacker's knowledge about each of the components ($k.i$)-($k.iv$).
Typically, the attacker is assumed to know the feature representation $\set X$ and the kind of learning algorithm $f$ (\eg, the fact that the classifier is linear, or it is a neural network with a given architecture, \etc), but neither the training data $\set D$ nor the classifier's (trained) parameters $\vct w$. The attacker is however assumed to be able to collect a surrogate data set $\hat{\set D}$\footnote{We use here the \emph{hat} symbol to denote limited knowledge of a given component.} from a similar source (ideally sampling from the same underlying data distribution), and potentially get feedback from the classifier about its decisions to provide labels for such data. This enables the attacker to estimate the parameters $\hat{\vct w}$ from $\hat{\set D}$, by training a \emph{surrogate classifier}.
We refer to this case as {LK attacks with Surrogate Data} (LK-SD), and denote it with $\vct \theta_{\rm LK-SD}=(\hat{\set D}, \set X, f, \hat{\vct w})$ .

We refer to the setting in which the attacker does not even know the kind of learning algorithm $f$ as {LK attacks with Surrogate Learners} (LK-SL), and denote it with $\vct \theta_{\rm LK-SL}=(\hat{\set D}, \set X, \hat{f}, \hat{\vct w})$.
LK-SL attacks also include the case in which the attacker knows the learning algorithm, but optimizing the attack samples against it may be not tractable or too complex. In this case, the attacker can also craft the attacks against a surrogate classifier and test them against the targeted one. 
This is a common procedure used also to evaluate the \emph{transferability} of attacks between learning algorithms, as firstly shown in~\cite{biggio13-ecml} and subsequently in~\cite{papernot17-asiaccs} for deep networks.

\myparagraph{Zero-Knowledge (ZK) Black-Box Attacks} 
Recent work has also claimed that machine learning can be threatened without any substantial knowledge of the feature space, the learning algorithm and the training data, if the attacker can query the system in a black-box manner and get feedback on the provided labels or confidence scores~\cite{tramer16-usenix,xu16-ndss,papernot17-asiaccs,chen17-aisec,dang17-ccs}.
This point deserves however some clarification.
First, the attacker knows (as any other potential user) that the classifier is designed to perform some task (\eg, object recognition in images, malware classification, \etc), and has to clearly have an idea of which potential transformations to apply to cause some feature changes, otherwise neither change can be inflicted to the output of the classification function, nor any useful information can be extracted from it. For example, if one attacks a malware detector based on dynamic analysis by injecting static code that will never be executed, there will be no impact at all on the classifier's decisions.
This means that, although the exact feature representation may be not known to the attacker, at least she knows (or has to get to know) which kind of features are used by the system (\eg, features based on static or dynamic analysis in malware detection). Thus, knowledge of the feature representation may be partial, but not completely absent. This is even more evident for deep networks trained on images, where the attacker knows that the input features \emph{are} the image pixels. 

Similar considerations hold for knowledge of the training data.
If the attacker knows that the classifier is used for a specific task, it is clear the she also knows which kind of data has been used to train it; for example, if a deep network aims to discriminate among classes of animals, then it is clear that it has been trained on images of such animals. Hence, also in this case the attacker effectively has some knowledge of the training data, even if not of the exact training samples.

We thus characterize this setting as $\vct \theta_{\rm ZK} = (\hat{\set D}, \hat{\set X}, \hat{f}, \hat{\vct w})$. Even if surrogate learners are not necessarily used here~\cite{tramer16-usenix,xu16-ndss,dang17-ccs,chen17-aisec}, as well as in pioneering work on black-box attacks against machine learning~\cite{lowd05,nelson12-jmlr}, one may anyway learn a surrogate classifier (potentially on a different feature representation) and check whether the crafted attack samples \emph{transfer} to the targeted classifier. Feedback from classifier's decisions on carefully-crafted query samples can then be used to refine the surrogate model, as in~\cite{papernot17-asiaccs}.
Although the problem of learning a surrogate model while minimizing the number of queries can be casted as an \emph{active learning} problem, to our knowledge well-established \emph{active learning} algorithms have not yet been compared against such recently-proposed approaches~\cite{papernot17-asiaccs}. 

\subsection{Attacker's Capability} \label{sect:cap}

This characteristic depends on the \emph{influence} that the attacker has on the input data, and on application-specific \emph{data manipulation constraints}.

\myparagraph{Attack Influence} It can be causative, if the attacker can manipulate both training and test data, or exploratory, if the attacker can only manipulate test data. These scenarios are more commonly known as \emph{poisoning} and \emph{evasion} attacks~\cite{barreno06-asiaccs,barreno10,huang11,biggio14-tkde,biggio13-ecml,biggio14-ijprai,biggio12-icml,biggio15-icml,mei15-aaai,biggio17-aisec}. 

\myparagraph{Data Manipulation Constraints} Another aspect related to the attacker's capability depends on the presence of application-specific constraints on data manipulation, \eg, to evade malware detection, malicious code has to be modified without compromising its intrusive functionality. 
This may be done against systems based on static code analysis, by injecting instructions or code that will never be executed~\cite{srndic14,biggio13-ecml,demontis17-tdsc,grosse17-esorics}.
These constraints can be generally accounted for in the definition of the optimal attack strategy by assuming that the initial attack samples $\set D_{c}$ can only be modified according to a space of possible modifications $\Phi(\set D_{c})$. In some cases, this space can also be mapped in terms of constraints on the feature values of the attack samples; \eg, by imposing that feature values corresponding to occurrences of some instructions in static malware detectors can only be incremented~\cite{srndic14,biggio13-ecml,demontis17-tdsc}.

\begin{figure}[t]
\begin{center}
\includegraphics[width=0.4\textwidth]{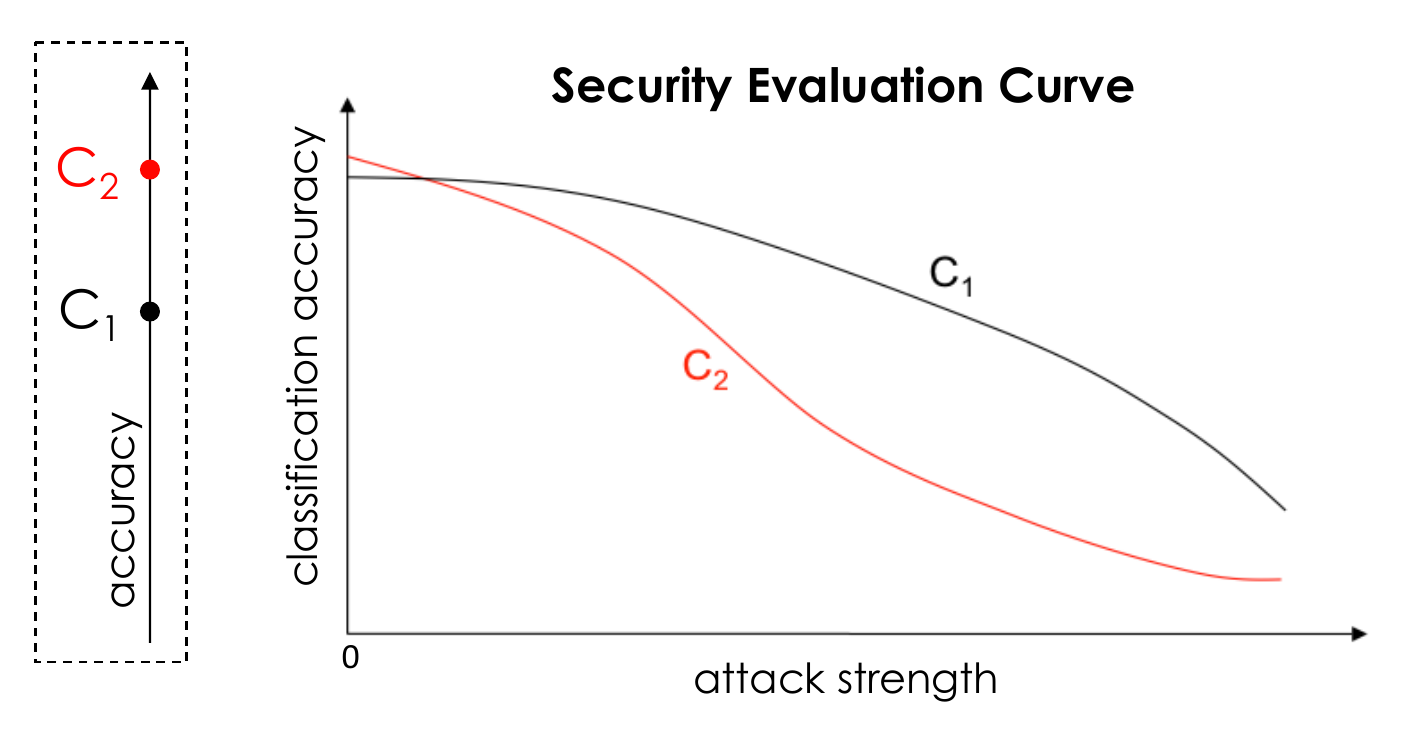}
\caption{Security evaluation curves for two hypothetical classifiers $C_1$ and $C_2$, inspired from the methodology proposed in~\cite{biggio14-tkde,biggio14-ijprai}. Based only on classification accuracy in the absence of attack, one may prefer $C_2$ to $C_1$. Simulating attacks of increasing \emph{strength} (\eg, by increasing the level of perturbation in input images) may however reveal that more accurate classifiers may be less robust to adversarial perturbations. Thus, one may finally prefer $C_1$ to $C_2$.}
\label{fig:sec}
\end{center}
\end{figure}

\subsection{Attack Strategy}
Given the attacker's knowledge $\vct \theta\in\Theta$ and a set of manipulated attack samples $\set D_{c}^{\prime} \in \Phi(\set D_{c})$, the attacker's goal can be defined in terms of an objective function $\set A (\set D_{c}^{\prime}, \vct \theta) \in \mathbb R$ which measures how effective the attacks $\set D_{c}^{\prime}$ are.
The optimal attack strategy can be thus given as: 
\begin{eqnarray}
\set D_{c}^{\star} \in \argmax_{\set D_{c}^{\prime} \in \Phi(\set D_{c})}  \set A(\set D_{c}^{\prime}, \vct \theta) 
\label{eq:optim}
\end{eqnarray}
We show in Sect.~\ref{sect:attacks} how this high-level formulation encompasses both evasion and poisoning attacks against \emph{supervised} learning algorithms, despite it has been used also to attack clustering~\cite{biggio13-aisec,biggio14-aisec,biggio14-spr}, and feature selection algorithms~\cite{biggio15-icml,zhang16-tcyb}.

\subsection{Security Evaluation Curves}

It is worth remarking here that, to provide a thorough security evaluation of learning algorithms, one should assess their performance not only under different assumptions on the attacker's knowledge, but also increasing the \emph{attack strength}, \ie, the attacker's capability $\Phi(\set D_c)$ of manipulating the input data. For example, this can be done by increasing the amount of perturbation used to craft evasion attacks, or the number of poisoning attack points injected into the training data. 
The resulting \emph{security evaluation curve}, conceptually represented in~Fig.~\ref{fig:sec}, shows the extent to which the performance of a learning algorithm drops more or less gracefully under attacks of increasing \emph{strength}.
This is crucial to enable a fairer comparison among different attack algorithms and defenses in the context of concrete application examples~\cite{biggio14-tkde,biggio14-ijprai}, as we will discuss in the remainder of this work.

\subsection{Summary of Attacks against Machine Learning}

\begin{figure}[t]
\begin{center}
\includegraphics[width=0.5\textwidth]{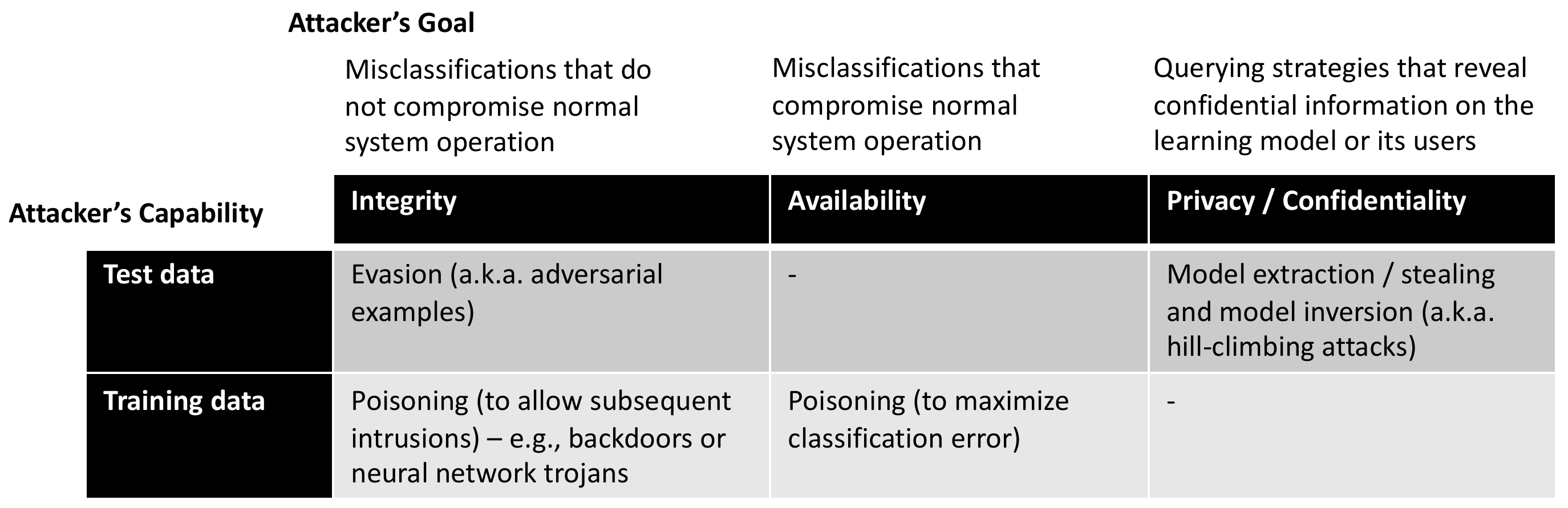}
\vspace{-5pt}
\caption{Categorization of attacks against machine learning based on our threat model.}
\vspace{-5pt}
\label{fig:ml-attacks}
\end{center}
\end{figure}

\begin{figure}[t]
\begin{center}
\includegraphics[width=0.45\textwidth]{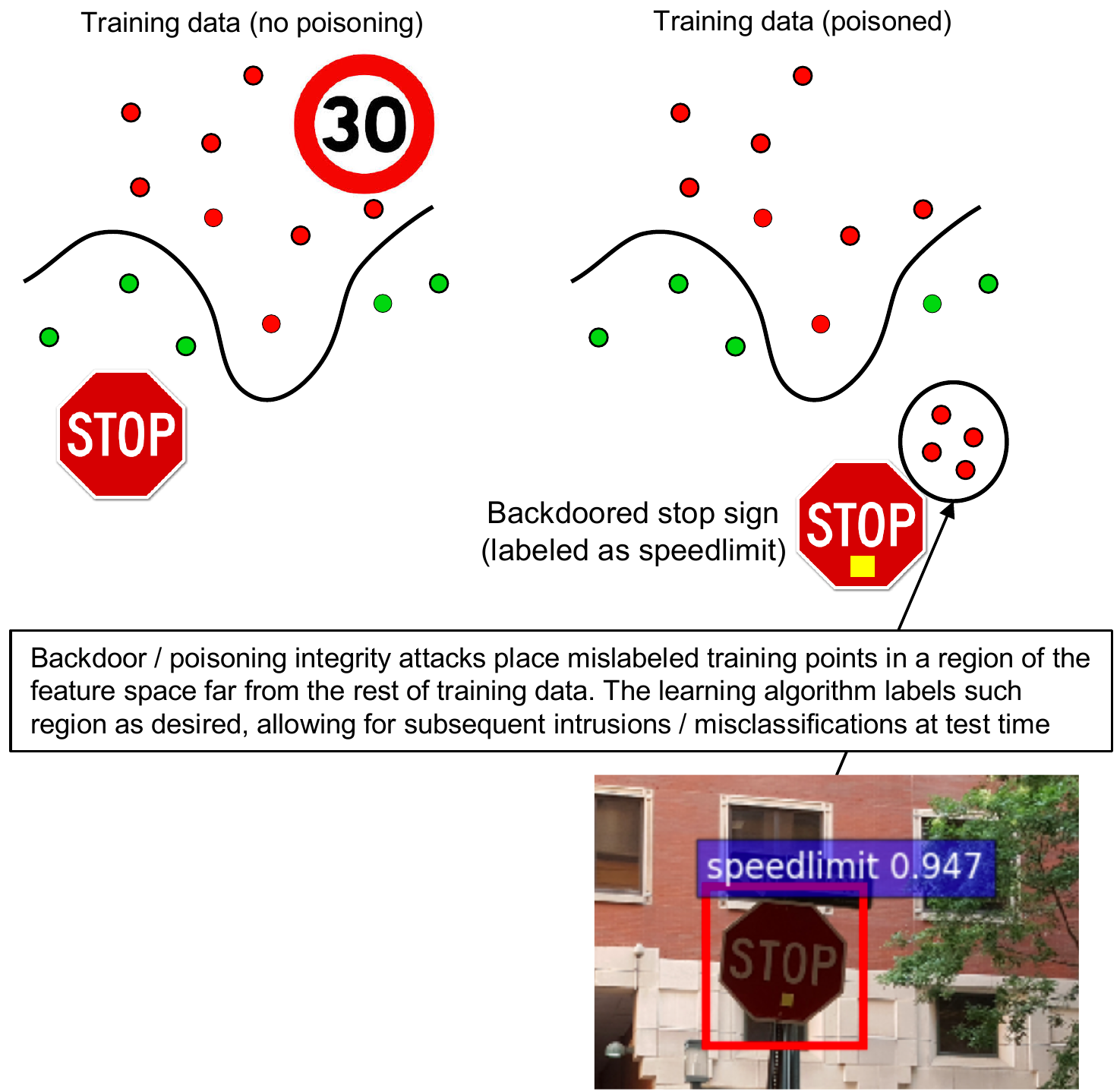}
\caption{Conceptual representation of the impact of poisoning integrity attacks (including backdoor and trojaning attacks) on the decision function of learning algorithms. The example, taken from~\cite{gu17}, shows a backdoored stop sign misclassified, as expected, as a speedlimit sign.}
\label{fig:backdoors}
\end{center}
\end{figure}

Before delving into the details of specific attacks, we provide in Fig.~\ref{fig:ml-attacks} a simplified categorization of the main attacks against machine-learning algorithms based on the aforementioned threat model; in particular, considering the attacker's goal and main capabilities. 
The most common attacks, as discussed before, include evasion and poisoning availability attacks (aimed to maximize the test error)~\cite{barreno06-asiaccs,barreno10,huang11,biggio14-tkde,biggio13-ecml,biggio14-ijprai,biggio12-icml,biggio15-icml,mei15-aaai,biggio17-aisec}. More recently, different kinds of poisoning integrity attacks (which manipulate the training data or the trained model to cause specific misclassifications, as defined in~\cite{biggio15-icml,biggio17-aisec}) against deep networks have been also studied under the name of \emph{backdoor} and \emph{trojaning} attacks~\cite{gu17,chen17}. These attacks maliciously manipulate pre-trained network models to create specific backdoor vulnerabilities. The corrupted models are then publicly released, to favor their adoption in proprietary systems (\eg, via fine-tuning or other transfer learning techniques).
When this happens, the attacker can activate the backdoors using specific input samples that are misclassified as desired. The underlying idea behind such attacks and their impact on the learning process is conceptually depicted in Fig.~\ref{fig:backdoors}. 
All the aforementioned attacks can be successfully staged under different levels of the attacker's knowledge.
When knowledge is limited, as in the gray-box and black-box cases, privacy or confidential attacks can be staged to gain further knowledge about the target classifier or its users. Although we do not thoroughly cover such attacks in detail here, we refer the reader to few practical examples of such attacks reported to date, including model-extraction attacks aimed to steal machine-learning models, and model-inversion and hill-climbing attacks against biometric systems used to steal the face and fingerprint templates of their users (or any other sensitive information)~\cite{biggio14-svm-chapter,fredrikson15-ccs,tramer16-usenix,adler05,galbally09,martinez11}.

\vspace{-5pt}
\section{Be Proactive: Simulating Attacks}
\label{sect:attacks}
\vspace{-5pt}

\begin{quote}
\begin{flushright}
\emph{``To know your enemy, you must become your enemy.''}\\
(Sun Tzu, The Art of War, 500 BC)
\end{flushright}
\end{quote}

We discuss here how to formalize test-time evasion and training-time poisoning attacks in terms of the optimization problem given in Eq.~\eqref{eq:optim}, and consistently with the threat model discussed in Sect.~\ref{sect:attack-framework}. Note that other attacks may also be described, generally, in terms of the aforementioned optimization problem, but we focus here only on those for which this connection is tighter.

\subsection{Evasion attacks} \label{sect:evasion-attacks}

Evasion attacks consist of manipulating input data to evade a trained classifier at test time. These include, \eg, manipulation of malware code to have the corresponding sample misclassified as legitimate, or manipulation of images to mislead object recognition.
We consider here the formulation reported in~\cite{melis17-vipar}, which extends our previous work~\cite{biggio13-ecml} from two-class to multiclass classifiers, by introducing \emph{error-generic} and \emph{error-specific} maximum-confidence evasion attacks. With reference to Eq.~\eqref{eq:optim}, the evasion attack samples $\set D^\prime_c$ can be optimized one at a time, independently, aiming to maximize the classifier's confidence associated to a wrong class. 
We will denote with $f_i(\vct x)$ the confidence score of the classifier on the sample $\vct x$ for class $i$.
These attacks can be optimized under different levels of attacker's knowledge through the use of surrogate classifiers, so we omit the distinction between $f_i(\vct x)$ and $\hat f_i(\vct x)$ below for notational convenience.  

\myparagraph{Error-generic Evasion Attacks} In this case, the attacker is interested in misleading classification, regardless of the output class predicted by the classifier. The problem can be thus formulated as:
\begin{eqnarray}
\label{eq:indiscriminate-1}
\max_{\vct x^{\prime}}  & & \set A( \vct x^\prime, \vct \theta) = \Omega(\vct x^\prime) = \max_{l \neq k} f_{l}(\vct x) - f_{k}(\vct x) \, ,\\
\label{eq:indiscriminate-2}
{\rm s.t. } && d(\vct x, \vct x^{\prime}) \leq d_{\rm max} \, , \; \, \vct x_{\rm lb} \preceq \vct x^{\prime} \preceq \vct x_{\rm ub} \, ,
\end{eqnarray}
where $f_{k}(\vct x)$ denotes the discriminant function associated to the true class $k$ of the source sample $\vct x$, and $\max_{l \neq k} f_{l}(\vct x)$ is the closest competing class (\ie, the one exhibiting the highest value of the discriminant function among the remaining classes).
The underlying idea behind this attack formulation, similarly to~\cite{moosavi16-deepfool}, is to ensure that the attack sample will be no longer classified correctly as a sample of class $k$, but rather misclassified as a sample of the closest candidate class.
The manipulation constraints $\Phi(\set D_c)$ are given in terms of: ($i$) a distance constraint $d(\vct x, \vct x^{\prime}) \leq d_{\rm max}$, which sets a bound on the maximum input perturbation between $\vct x$ (\ie, the input sample) and the corresponding modified adversarial example $\vct x^{\prime}$; and ($ii$) a box constraint $\vct x_{\rm lb} \preceq \vct x^{\prime} \preceq \vct x_{\rm ub}$ (where $\vct u \preceq \vct v$ means that each element of $\vct u$ has to be not greater than the corresponding element in $\vct v$), which bounds the values of the attack sample $\vct x^\prime$.

For images, the former constraint is used to implement either \emph{dense} or \emph{sparse} evasion attacks~\cite{demontis16-spr,russu16-aisec,melis17-vipar}. Normally, the $\ell_{2}$ and the $\ell_{\infty}$ distances between pixel values are used to cause an indistinguishable image blurring effect (by slightly manipulating all pixels).
Conversely, the $\ell_{1}$ distance corresponds to a sparse attack in which only few pixels are significantly manipulated, yielding a salt-and-pepper noise effect on the image~\cite{demontis16-spr,russu16-aisec}. 
In the image domain, the box constraint can be used to bound each pixel value between $0$ and $255$, or to ensure manipulation of only a specific region of the image.
For example, if some pixels should not be manipulated, one can set the corresponding values of $\vct x_{\rm lb}$ and $\vct x_{\rm ub}$ equal to those of $\vct x$.
This is of interest to create real-world adversarial examples, as it avoids the manipulation of background pixels which do not belong to the object of interest~\cite{melis17-vipar,sharif16-ccs}.
Similar constraints have been applied also for evading learning-based malware detectors~\cite{biggio13-ecml,demontis16-spr,russu16-aisec,srndic14,demontis17-tdsc}.

\begin{figure}[t]
\centering
\includegraphics[width=0.2\textwidth]{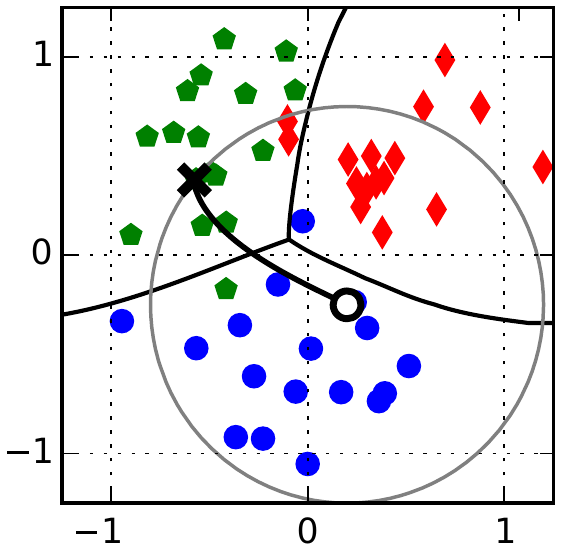}
\includegraphics[width=0.2\textwidth]{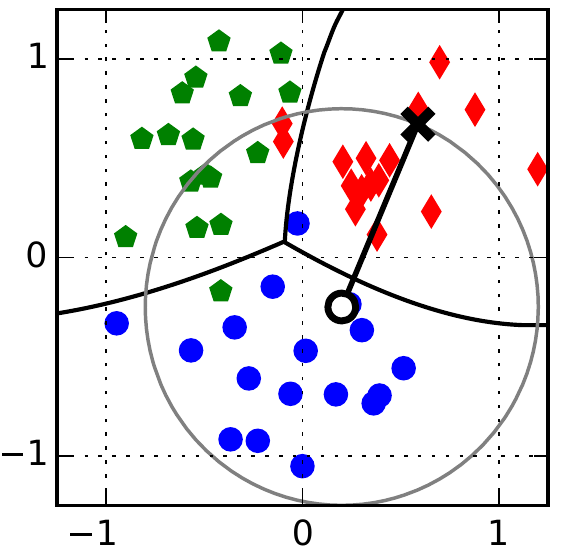}
\caption{Examples of error-specific (\emph{left}) and error-generic (\emph{right}) evasion, as reported in~\cite{melis17-vipar}. Decision boundaries among the three classes (blue, red and green points) are shown as black lines. In the error-specific case, the initial (blue) sample is shifted towards the green class (selected as target). In the error-generic case, instead, it is shifted towards the red class, as it is the closest class to the initial sample. The gray circle represents the feasible domain, given as an upper bound on the $\ell_2$ distance between the initial and the manipulated attack sample.}
\label{fig:eva-targeted-indiscriminate}
\end{figure}

\myparagraph{Error-specific Evasion Attacks} In the \emph{error-specific} setting, the attacker aims to mislead classification, but she requires the adversarial examples to be misclassified as a specific class.
The problem is formulated similarly to error-generic evasion (Eqs.~\ref{eq:indiscriminate-1}-\ref{eq:indiscriminate-2}), with the only differences that: ($i$) the objective function $\set A( \vct x^\prime, \vct \theta) = - \Omega(\vct x^\prime)$ has opposite sign; and ($ii$) $f_{k}$ denotes the discriminant function associated to the targeted class, \ie, the class which the adversarial example should be (wrongly) assigned to.
The rationale in this case is to maximize the confidence assigned to the wrong target class $f_k$, while minimizing the probability of correct classification~\cite{melis17-vipar,moosavi16-deepfool}.

\myparagraph{Attack Algorithm} The two evasion settings are conceptually depicted in Fig.~\ref{fig:eva-targeted-indiscriminate}. Both can be solved through a straightforward gradient-based attack, for differentiable learning algorithms (including neural networks, SVMs with differentiable kernels, \etc)~\cite{biggio13-ecml,melis17-vipar}. Non-differentiable learning algorithms, like decision trees and random forests, can be attacked with more complex strategies~\cite{kantchelian16-icml} or using the same algorithm against a differentiable surrogate learner~\cite{russu16-aisec}.

\begin{figure*}[t]
\centering
\includegraphics[height=0.2\textwidth]{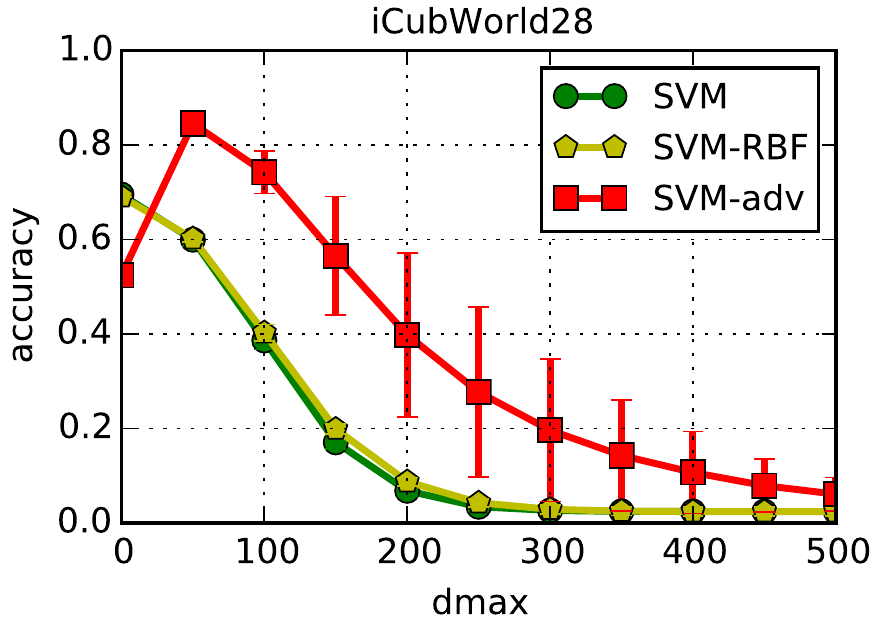}
\includegraphics[height=0.2\textwidth]{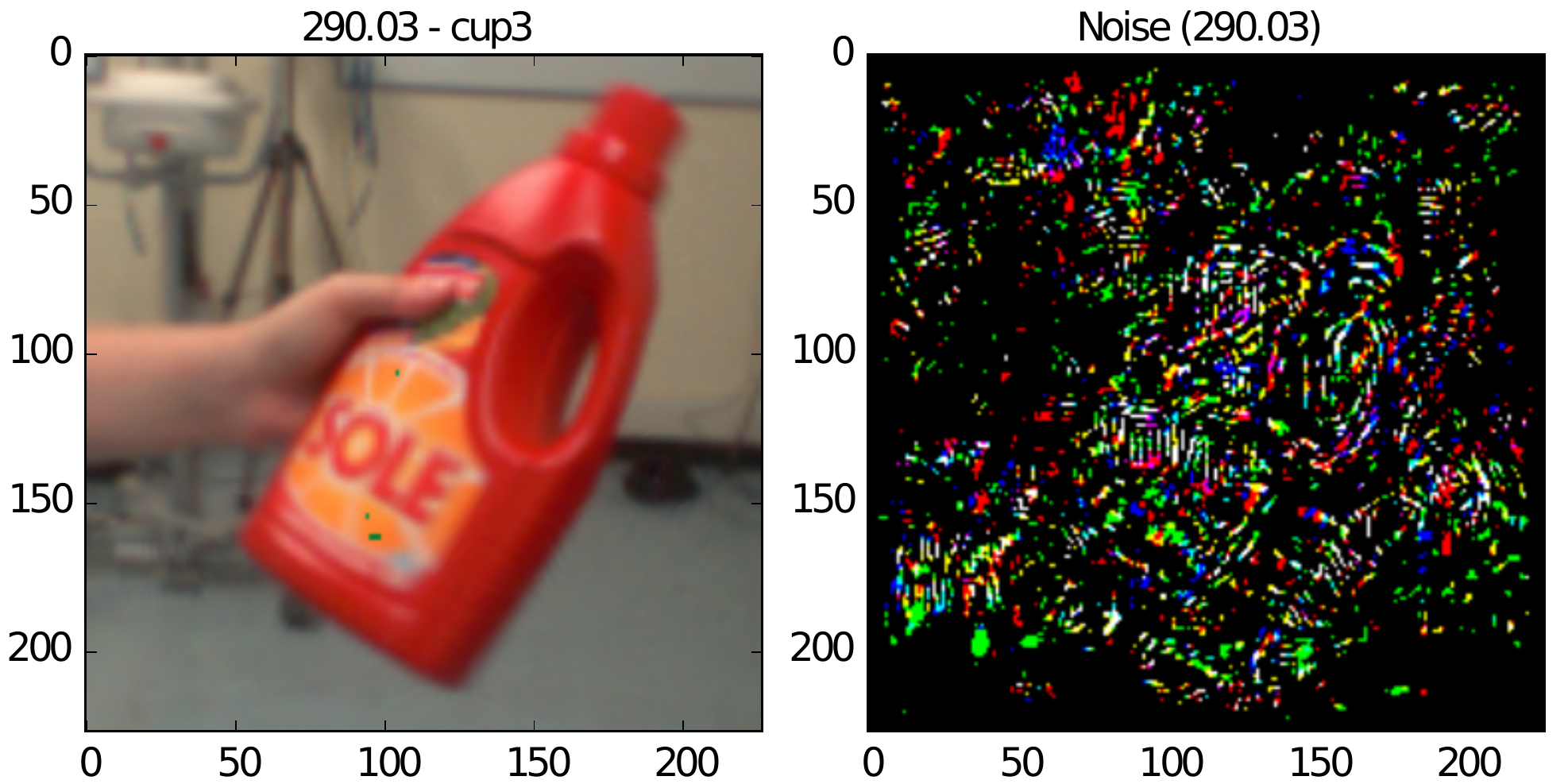}\\
\includegraphics[height=0.2\textwidth]{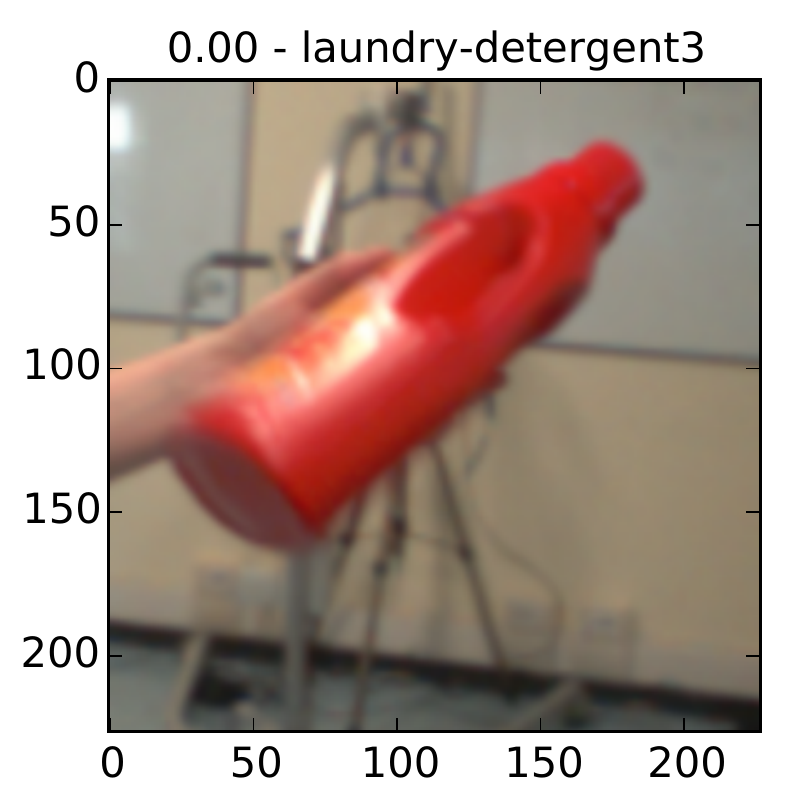}
\includegraphics[height=0.2\textwidth]{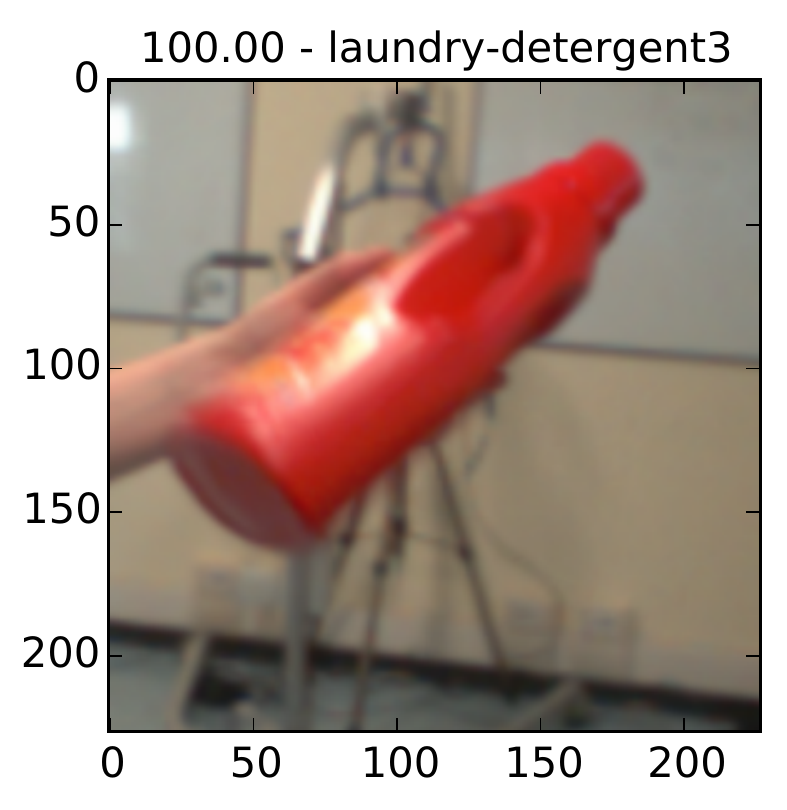}
\includegraphics[height=0.2\textwidth]{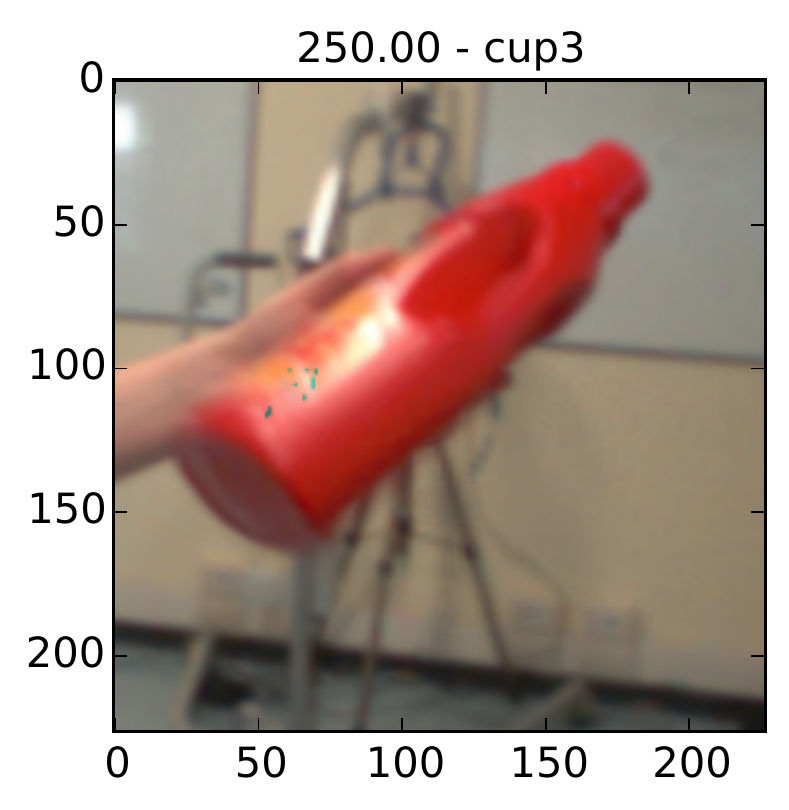}
\includegraphics[height=0.2\textwidth]{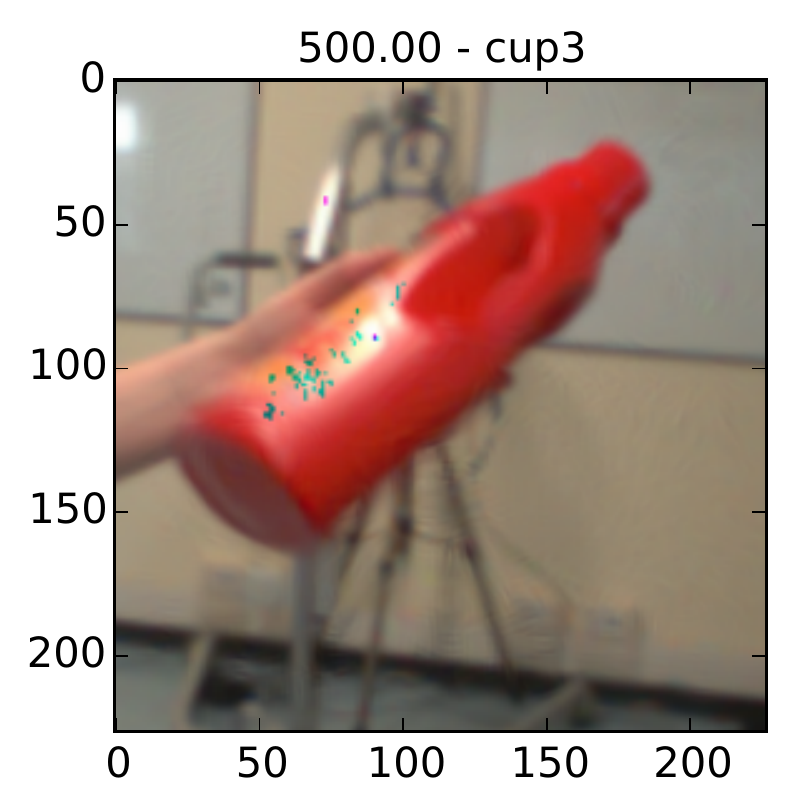}
\vspace{-10pt}
\caption{Error-specific evasion results from~\cite{melis17-vipar}.
\emph{Top row:} Security evaluation curves reporting accuracy of the given classifiers against an increasing $\ell_2$ input perturbation. The right-hand side plots depict a laundry detergent misclassified as a cup when applying the minimum input perturbation required for misclassification, along with the corresponding magnified noise mask. \emph{Bottom row:} Images of the laundry detergent perturbed with an increasing level of noise. The manipulations are only barely visible for perturbation values higher than $150$-$200$ (recall however that these values depend on the image size, as the $\ell_2$ distance).
}
\label{fig:evasion-example}
\vspace{-5pt}
\end{figure*}

\subsubsection{Application Example} \label{sect:evasion-app}
We report here an excerpt of the results from our recent work~\cite{melis17-vipar}, where we have constructed adversarial examples aimed to fool the robot-vision system of the iCub humanoid.\footnote{\url{http://www.icub.org}}
This system uses a deep network to compute a set of deep features from input images (\ie, by extracting the output of the penultimate layer of the network), and then learns a multiclass classifier on this representation for recognizing $28$ different objects, including cups, detergents, hair sprayers, \etc 
The results for error-specific evasion (averaged on different target classes) are reported in Fig.~\ref{fig:evasion-example}, along with some examples of perturbed input images at different levels. We trained multiclass linear SVMs (SVM), SVMs with the RBF kernel (SVM-RBF), and also a simple defense mechanism against adversarial examples based on rejecting samples that are sufficiently far (in deep space) from known training instances (SVM-adv). This will be discussed more in detail in Sect.~\ref{sect:static-defenses} (see also Fig.~\ref{fig:svm-rbf-reject} for a conceptual representation of this defense mechanism).
The security evaluation curves in Fig.~\ref{fig:evasion-example} show how classification accuracy decreases against an increasing $\ell_2$ maximum admissible perturbation $d_{\rm max}$. Notably, the rejection mechanism of SVM-adv is only effective for low input perturbations (at the cost of some additional misclassifications in the absence of attack). For higher perturbation levels, the deep features of the manipulated attacks become indistinguishable to those of the samples of the targeted class, although the input image is still far from resembling a different object. This phenomenon is connected to the instability of the deep representation learned by the underlying deep network. We refer the reader to \cite{melis17-vipar} for further details, and to~\cite{athalye18-iclr} (and references therein) for the problem of generating adversarial examples in the physical world.

\subsubsection{Historical Remarks}
We conclude this section with some historical remarks on evasion attacks, with the goal of providing a better understanding of the connections with recent work on adversarial examples and the security of deep learning. 

Evasion attacks have a long tradition. As mentioned in Sect.~\ref{sect:intro}, back in 2004-2006, work in~\cite{wittel04,lowd05,lowd05-ceas,fogla06} reported preliminary attempts in evading statistical anti-spam filters and malware detectors with ad-hoc evasion strategies. The very first evasion attacks against linear classifiers were systematized in the same period in~\cite{dalvi04,lowd05,lowd05-ceas}, always considering spam filtering as a running example.
The underlying idea was to manipulate the content of spam emails by obfuscating \emph{bad} words and/or adding \emph{good} words. To reduce the number of manipulated words in each spam, and preserve message readability, the idea was to modify first words which were assigned the highest absolute weight values by the linear text classifier.
Heuristic countermeasures were also proposed before 2010~\cite{globerson06-icml,kolcz09,biggio10-ijmlc}, based on the intuition of learning linear classifiers with more \emph{uniform} feature weights, to require the attacker to modify more words to get her spam misclassified.
To summarize, the vulnerability of linear classifiers to evasion attacks was a known problem even prior to 2010, and simple, heuristic countermeasures were already under development.
Meanwhile, Barreno et al. (see~\cite{barreno06-asiaccs,barreno10} and references therein) were providing an initial overview of the vulnerabilities of machine learning from a more general perspective, highlighting the need for \emph{adversarial} machine learning, \ie, to develop learning algorithms that explicitly account for the presence of the attacker~\cite{huang11}.

At that time, the idea that nonlinear classifiers could be more robust than linear ones against evasion was also becoming popular. In 2013, \v{S}rndi\'{c} and Laskov~\cite{srndic13-ndss} proposed a learning-based PDF malware detector, and attacked it to test its vulnerability to evasion. They reported that:
\vspace{-4pt}
\begin{quote}
\emph{The most aggressive evasion strategy we could conceive was successful for only 0.025\% of malicious examples tested against a nonlinear SVM classifier with the RBF kernel [...] we do not have a rigorous mathematical explanation for such a surprising robustness. Our intuition suggests that [...] the space of true features is hidden behind a complex nonlinear transformation which is mathematically hard to invert. [...] hence, the robustness of the RBF classifier must be rooted in its nonlinear transformation.}
\end{quote}
\vspace{-4pt}
Today we know that this hypothesis about the robustness of nonlinear classifiers is wrong.
The fact that a system could be more secure against an attack not specifically targeted against it does not provide any further meaningful information about its security to more powerful worst-case attacks.
Different systems (and algorithms) should be tested under the same (worst-case) assumptions on the underlying threat model. In particular, it is not difficult to see that the attack developed in that work was somehow crafted to evade linear classifiers, but not sufficiently complex to fool nonlinear ones.

While reading that work, it was thus natural to ask ourselves: ``what if the attack is carefully-crafted against nonlinear classifiers, instead? How can we invert such complex nonlinear transformation to understand which features are more relevant to the classification of a sample, and change them?''
The answer was readily available: the gradient of the classification function is exactly what specifies the direction of maximum variation of the function with respect to the input features.
Thus, we decided to formulate the evasion of a nonlinear classifier similarly to what we did in~\cite{biggio10-ijmlc} for linear classifiers, in terms of an optimization problem that minimizes the discriminant function $f(\vct x)$ such that $\vct x$ is misclassified as legitimate with maximum confidence, under a maximum amount of possible changes to its feature vector.

In a subsequent paper~\cite{biggio13-ecml}, we implemented the aforementioned strategy and showed how to evade nonlinear SVMs and neural networks through a straightforward gradient-descent attack algorithm. In the same work, we also reported the first ``adversarial examples'' on MNIST handwritten digit data against nonlinear learning algorithms.
We furthermore showed that, when the attacker does not have perfect knowledge of the targeted classifier, a surrogate classifier can be learned on surrogate training data, and used to craft the attack samples which then \emph{transfer} with high probability to the targeted model. This was also the first experiment showing that adversarial examples can be transferred, at least in a gray-box setting (training the same algorithm on different data). Notably, \v{S}rndi\'{c} and Laskov~\cite{srndic14} subsequently exploited this attack to show that PDF malware detectors based on nonlinear learning algorithms were also vulnerable to evasion, conversely to what they supposed in~\cite{srndic13-ndss}.

More recently, we have also exploited the theoretical findings  in~\cite{xu09}, which connect regularization and robustness in kernel-based classifiers, to provide a theoretically-sound countermeasure for linear classifiers against evasion attacks~\cite{demontis16-spr,demontis17-tdsc}. These recent developments have enabled a deeper understanding on how to defend against evasion attacks in spam filtering and malware detection, also clarifying (in a formal manner) the intuitive idea of \emph{uniform} feature weights only heuristically provided in~\cite{kolcz09,biggio10-ijmlc}. In particular, we have recently shown how a proper, theoretically-grounded regularization scheme can significantly outperform heuristic approaches in these contexts~\cite{demontis16-spr,demontis17-tdsc}.

\myparagraph{Security of Deep Learning} In 2014-2015, Szegedy \etal~\cite{szegedy14-iclr} and subsequent work~\cite{goodfellow15-iclr,nguyen15-cvpr,moosavi16-deepfool} showed that deep networks can be fooled by well-crafted, minimally-perturbed input images at test time, called \emph{adversarial examples}.
These samples are obtained by minimizing the distance of the adversarial sample $\vct x^\prime$ to the corresponding source sample $\vct x$ under the constraint that the predicted label is different, \ie, $\min_{\vct x^\prime} d(\vct x, \vct x^\prime) $ s.t. $f(\vct x) \neq f(\vct x^\prime)$.
Interestingly, the parallel discovery of such gradient-based adversarial perturbations by Szegedy \etal~\cite{szegedy14-iclr} and Biggio \etal~\cite{biggio13-ecml} started from different premises, as also explained by Ian Goodfellow in one of his popular talks.\footnote{Available at: \url{https://youtu.be/CIfsB_EYsVI?t=6m02s}} 
While we were investigating how to evade detection by learning algorithms in security-related tasks with a clear adversarial nature (like spam and malware detection)~\cite{biggio13-ecml}, Szegedy \etal~\cite{szegedy14-iclr} were trying to interpret and visualize the salient characteristics learned by deep networks.
To this end, they started looking for minimal changes to images in the input space that cause misclassifications. They expected to see significant changes to the background of the image or to the structure and aspect of the depicted objects, while it turned out that, quite surprisingly, such modifications were almost imperceptible to the human eye.
This discovery has raised an enormous interest in both the computer vision and security communities which, since then, have started proposing novel security assessment methodologies, attacks and countermeasures to mitigate this threat, independently re-discovering many other aspects and phenomena that had been already known, to some extent, in the area of adversarial machine learning. 
An example is given by the use of surrogate classifiers with smoother decision functions~\cite{papernot17-asiaccs} to attack models that either mask gradients or are not end-to-end differentiable~\cite{papernot16-distill}.
One of such defense mechanisms, known as \emph{distillation}~\cite{papernot16-distill}, has indeed shown to be vulnerable to attacks based on surrogate classifiers~\cite{papernot17-asiaccs}, essentially leveraging the idea behind limited-knowledge evasion attacks that we first discussed in~\cite{biggio13-ecml}. Iterative attacks based on projected gradients have also been independently re-discovered~\cite{madry18-iclr}, despite they had been used before against other classifiers~\cite{biggio13-ecml,srndic14,melis17-vipar}.

\begin{figure*}[!t]
\begin{center}
\includegraphics[width=0.95\textwidth]{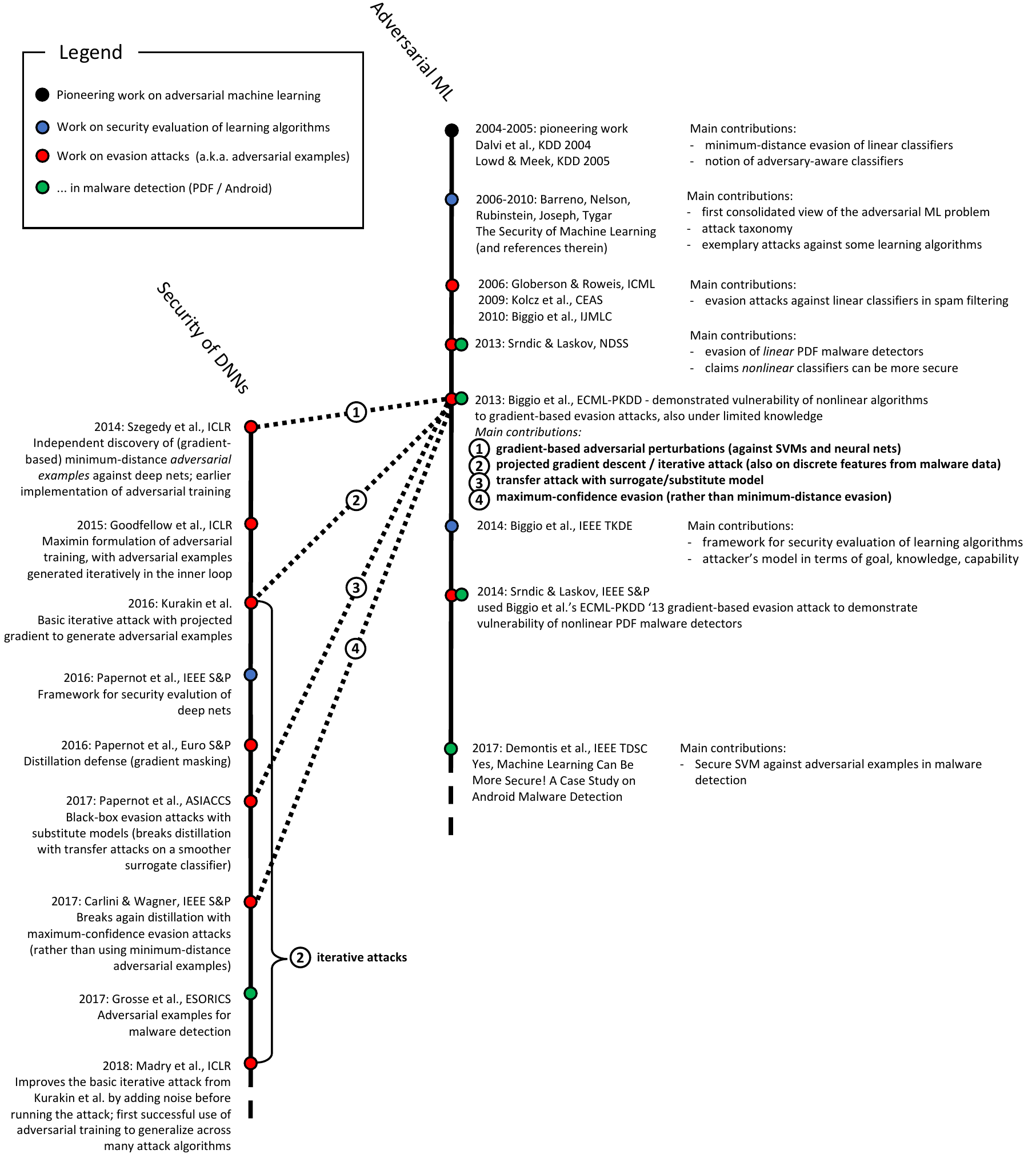}
\caption{Timeline of evasion attacks (\ie, adversarial examples) in adversarial machine learning, compared to work on the security of deep networks. Related work is highlighted with markers of the same color (as reported in the legend) and connected with dashed lines to highlight independent (but related) findings. The date of publication refers to publication on peer-reviewed conferences and journals.}
\label{fig:timeline}
\end{center}
\end{figure*}

\myparagraph{Misconceptions on Evasion Attacks} The main misconception that is worth highlighting here is that \emph{adversarial examples should be minimally perturbed}. 
The motivation of this misconception is easy to explain. The notion of adversarial examples was initially introduced to analyze the instability of deep networks~\cite{szegedy14-iclr}, \ie, their sensitivity to minimal input perturbations; the goal of the initial work on adversarial examples was not to to perform a detailed security assessment of a machine-learning algorithm using security evaluation curves (Fig.~\ref{fig:evasion-example}).
Normally, as already discussed in this paper and also in our previous work~\cite{biggio13-ecml,biggio14-tkde}, for the purpose of thoroughly assessing the security of a learning algorithm under attack, given a feasible space of modifications to the input data, it is more reasonable to assume that the attacker will aim to maximize the classifier's confidence on the desired output class, rather than only minimally perturbing the attack samples (cf. Eqs.~\ref{eq:indiscriminate-1}-\ref{eq:indiscriminate-2}).
For this reason, while minimally-perturbed adversarial examples can be used to analyze the sensitivity of a learning algorithm, maximum-confidence attacks are more suitable for a thorough security assessment of learning algorithms under attack. In particular, the use of the security evaluation curves described above  gives us a clearer understanding of the security of a learning algorithm under more powerful attacks. By increasing the attack strength (\ie, the maximum amount of perturbation applicable to the input data), one can draw a complete security evaluation curve, reporting the evasion rate for each value of attack strength. This ensures us that, \eg, if the noise applied to the input data is not larger than $\epsilon$, then the classification performance should not drop more than $\delta$. Conversely, using minimally-perturbed adversarial examples one can only provide guarantees against an average level of perturbation (rather than a worst-case bound).  
The fact that maximum- or high-confidence attacks are better suited to the task of security evaluation of learning algorithms (as well as to improve transferability across different models) is also witnessed by the work by Carlini and Wagner~\cite{carlini17-sp,carlini17-aisec} and follow-up work in~\cite{athalye18,dong18-cvpr}. In that work, the authors exploited a similar idea to show that several recent defenses proposed against minimally-perturbed adversarial examples are vulnerable to maximum-confidence ones, using a stronger attack similar to those proposed in our earlier work, and discussed in Sect.~\ref{sect:evasion-attacks}~\cite{biggio13-ecml,biggio14-tkde}.
Even in the domain of malware detection, adversarial examples seem to be a novel threat~\cite{grosse17-esorics}, while the vulnerability of learning-based malware detectors to evasion is clearly a consolidated issue~\cite{biggio13-ecml,srndic14,demontis17-tdsc}.
Other interesting avenues to provide reliable guarantees on the security of neural networks include \emph{formal verification}~\cite{huang17-cav} and evaluation methods inspired from software testing~\cite{pei17}.

\myparagraph{Timeline of Evasion Attacks} To summarize, while the security of deep networks has received considerable attention from different research communities only recently, it is worth remarking that several related problems and solutions had been already considered prior to 2014 in the field of adversarial machine learning. Maximum-confidence evasion attacks and surrogate models are just two examples of similar findings in both areas of research. We compactly and conceptually highlight these connections in the timeline reported in Fig.~\ref{fig:timeline}.\footnote{An online version of the timeline is also available at: \url{https://sec-ml.pluribus-one.it}, along with a web application that allows one to generate adversarial examples and evaluate if they are able to evade detection (evasion attacks).}

\subsection{Poisoning Attacks}

\begin{figure*}[t]
\centering
\includegraphics[width=0.8\textwidth]{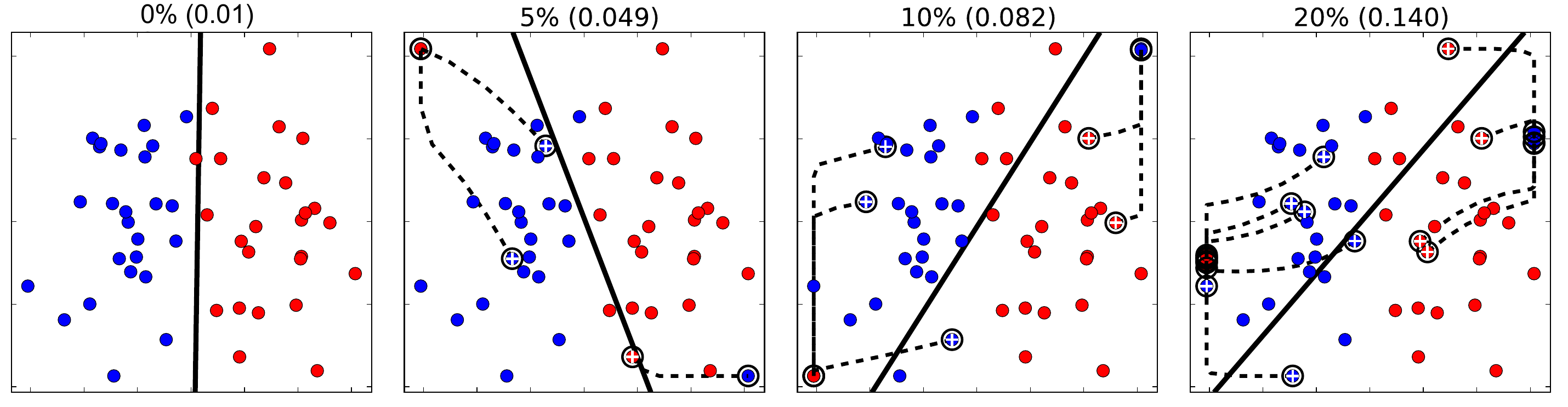}
\caption{Conceptual example of how poisoning attacks compromise a linear classifier~\cite{biggio12-icml,biggio15-icml}. Each plot shows the training data (red and blue points) and the corresponding trained classifier (black solid line). The fraction of poisoning points injected into the training set is reported on top of each plot, along with the test error (in parentheses) of the poisoned classifier. Poisoning points are optimized through a gradient-based attack algorithm. They are initialized by cloning the training points denoted with white crosses and flipping their label. The gradient trajectories (black dashed lines) are then followed up to some local optima to obtain the final poisoning points (highlighted with black circles).}
\label{fig:poisoning-ex}
\end{figure*}

Poisoning attacks aim to increase the number of misclassified samples at test time by injecting a small fraction of poisoning samples into the training data.
These attacks, conversely to evasion, are staged at the training phase.
A conceptual example of how poisoning works is given in Fig.~\ref{fig:poisoning-ex}. As for evasion attacks, we discuss here error-generic and error-specific poisoning attacks in a PK white-box setting, given that the extension to gray-box and black-box settings is trivial through the use of surrogate learners~\cite{biggio17-aisec}.

\myparagraph{Error-Generic Poisoning Attacks} In this case, the attacker aims to cause a \emph{denial of service}, by inducing as many misclassifications as possible (regardless of the classes in which they occur).
Poisoning attacks are generally formulated as bilevel optimization problems, in which the outer optimization maximizes the attacker's objective $\set A$ (typically, a loss function $L$ computed on untainted data), while the inner optimization amounts to learning the classifier on the poisoned training data~\cite{biggio12-icml,biggio15-icml,mei15-aaai}.
This can be made explicit by rewriting Eq.~\eqref{eq:optim} as:
\begin{align}
  \label{eqObjective1} \set D_c^{\star} \in & \argmax_{\set D_{c}^{\prime} \in \Phi(\set D_{c})} && \set A(\set D_{c}^{\prime}, \vct \theta) = L ({\set D}_{\rm val}, {\vct w^\star}) \, ,  \\
 \label{eqObjective2} &{\rm s.t.}   && {\vct w^\star} \in \argmin_{\vct w^{\prime} \in \set W} \set L({\set D}_{\rm tr} \cup \set D_{c}^{\prime}, \vct w^{\prime}) \, ,
\end{align}
where ${\set D}_{\rm tr}$ and ${\set D}_{\rm val}$ are two data sets available to the attacker. The former, along with the poisoning attack samples $\set D_{c}^{\prime}$, is used to train the learner on poisoned data, while the latter is used to evaluate its performance on untainted data, through the loss function $L ({\set D}_{\rm val}, {\vct w^\star})$. 
Notably, the objective function implicitly depends on $\set D_{c}^{\prime}$ through the parameters ${\vct w^\star}$ of the poisoned classifier. 

\myparagraph{Error-Specific Poisoning Attacks} In this setting the attacker aims to cause specific misclassifications. While the problem remains that given by Eqs.~\eqref{eqObjective1}-\eqref{eqObjective2}, the objective is redefined as $\set A(\set D_{c}^{\prime}, \vct \theta) = - L ({\set D}^{\prime}_{\rm val}, {\vct w^\star})$.
The set ${\set D}^{\prime}_{\rm val}$ contains the same samples as ${\set D}_{\rm val}$, but their labels are chosen by the attacker according to the desired misclassifications. The objective $L$ is then taken with opposite sign as the attacker effectively aims to \emph{minimize} the loss on her desired labels~\cite{biggio17-aisec}. 

\myparagraph{Attack Algorithm} A common trick used to solve the given bilevel optimization problems is to replace the inner optimization by its equilibrium conditions~\cite{biggio12-icml,biggio15-icml,mei15-aaai,biggio17-aisec}. This enables gradient computation in closed form and, thus, similarly to the evasion case, the derivation of gradient-based attacks (although gradient-based poisoning is much more computationally demanding, as it requires retraining the classifier iteratively on the modified attack samples).
In the case of deep networks, this approach is not practical due to computational complexity and instability of the closed-form gradients. 
To tackle this issue, we have recently proposed a more efficient technique, named \emph{back-gradient poisoning}. It relies on automatic differentiation and on reversing the learning procedure to compute the gradient of interest (see~\cite{biggio17-aisec} for further details).

\subsubsection{Application Example}

\begin{figure}[t]
\centering
\includegraphics[width=0.35\textwidth]{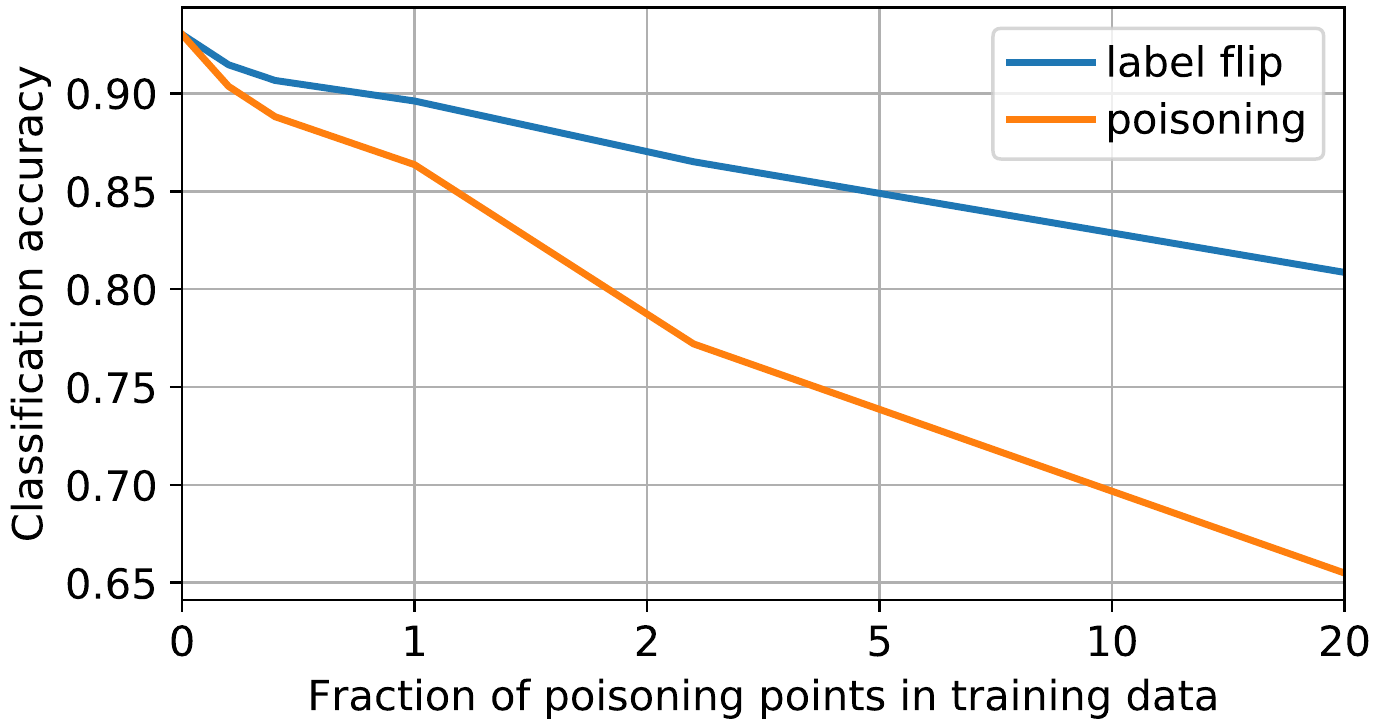}\\ \vspace{5pt}
\includegraphics[width=0.16\textwidth]{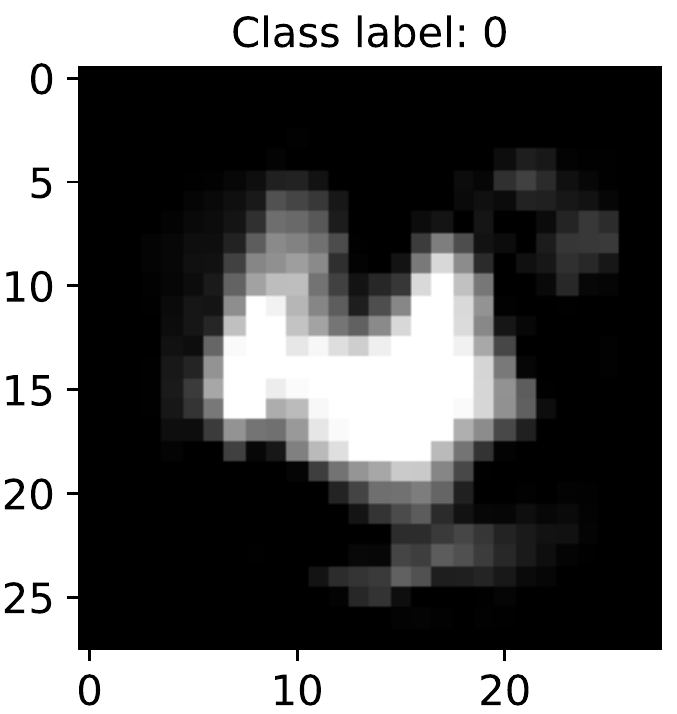}
\includegraphics[width=0.16\textwidth]{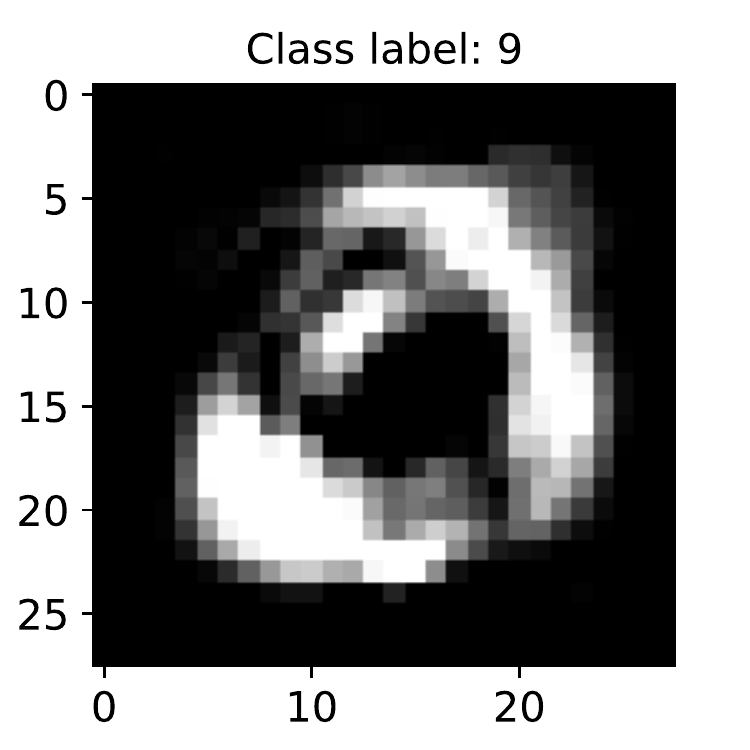}
\caption{Security evaluation curve of a softmax classifier trained on the MNIST digits $0$, $4$, and $9$, against back-gradient poisoning and random label flips (baseline comparison). Examples of adversarial training digits generated by back-gradient poisoning are shown on the right.}
\label{fig:poisoning}
\end{figure}

We report here an exemplary poisoning attack against a multiclass softmax classifier (logistic regression) trained on MNIST handwritten digits belonging to class $0$, $4$, and $9$. We consider error-generic poisoning, using $200$ (clean) training samples and $2000$ validation and test samples.
Results of back-gradient poisoning compared to randomly-injected training points with wrong class labels (random label flips) are reported in Fig.~\ref{fig:poisoning}, along with some \emph{adversarial training examples} generated by our back-gradient poisoning algorithm. 

\subsubsection{Historical Remarks}

To our knowledge, the earliest poisoning attacks date back to 2006-2010~\cite{newsome06,barreno06-asiaccs,nelson08,rubinstein09,kloft10}. Newsome et al.~\cite{newsome06} devised an attack to mislead signature generation for malware detection; Nelson~\etal~\cite{nelson08} showed that spam filters can be compromised to misclassify legitimate email as spam, by learning spam emails containing \emph{good} words during training; and Rubinstein~\etal~\cite{rubinstein09} showed how to poison an anomaly detector trained on network traffic through injection of \emph{chaff} traffic.
In the meanwhile, exemplary attacks against learning-based centroid anomaly detectors where also demonstrated~\cite{barreno06-asiaccs,kloft10,kloft12b}.
Using a similar formalization, we have also recently showed poisoning attacks against biometric systems~\cite{biggio15-spmag}.
This background paved the way to subsequent work that formalized poisoning attacks against more complex learning algorithms (including SVMs, ridge regression, and LASSO) as bilevel optimization problems~\cite{biggio12-icml,biggio15-icml,mei15-aaai}.
Recently, preliminary attempts towards poisoning deep networks have also been reported, showing the first \emph{adversarial training examples} against deep learners~\cite{biggio17-aisec,koh17-icml}.

It is worth finally remarking that poisoning attacks against machine learning should not be considered an academic exercise in vitro. Microsoft \emph{Tay}, a chatbot designed to talk to youngsters in Twitter, was shut down after only 16 hours, as it started raising racist and offensive comments after being poisoned.\footnote{\url{https://www.wired.com/2017/02/keep-ai-turning-racist-monster}} Its artificial intelligence was designed to mimic the behavior of humans, but not to recognize potential misleading behaviors. Kaspersky Lab, a leading antivirus company, has been accused of poisoning competing antivirus products through the injection of false positive examples into VirusTotal,\footnote{\url{http://virustotal.com}} although it is worth saying that they denied any wrongdoing, and blamed for spreading false rumors. 
Another avenue for poisoning arises from the fact that shared, big and open data sets are commonly used to train machine-learning algorithms. The case of ImageNet for object recognition is paradigmatic. In fact, people typically reuse these large-scale deep networks as feature extractors inside their pattern recognition tools. Imagine what may happen if someone could poison these data ``reservoirs'': many data-driven products and services could experience security and privacy issues, economic losses, with legal and ethical implications.

\vspace{-5pt}
\section{Protect Yourself: Security Measures for Learning Algorithms}
\label{sect:defenses}
\vspace{-5pt}

\begin{quote}
\begin{flushright}
\emph{``What is the rule? The rule is protect yourself at all times.''}\\
(from the movie \emph{Million dollar baby}, 2004)
\end{flushright}
\end{quote}

\begin{figure*}[t]
\begin{center}
\includegraphics[width=0.8\textwidth]{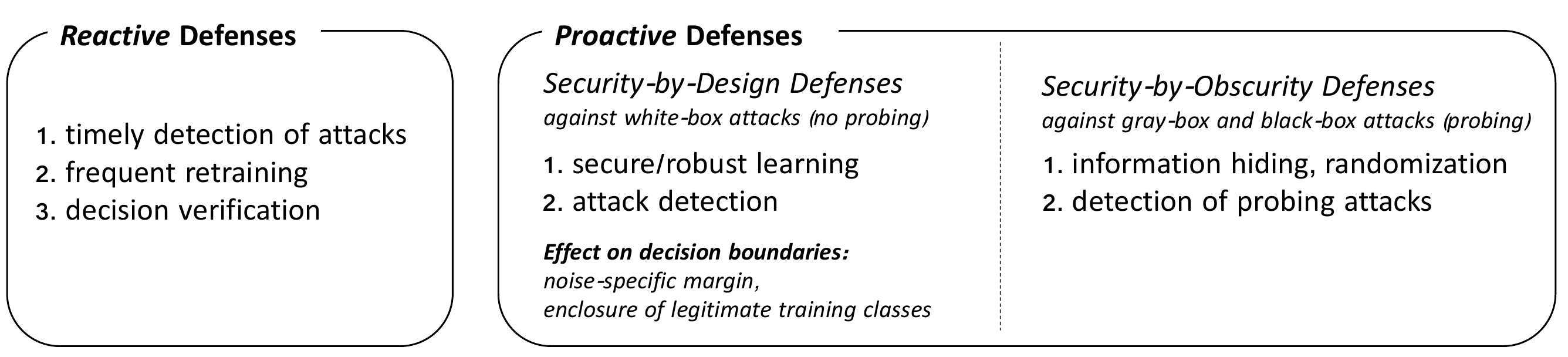}
\vspace{-10pt}
\caption{Schematic categorization of the defense techniques discussed in Sect.~\ref{sect:defenses}.}
\vspace{-15pt}
\label{fig:defenses}
\end{center}
\end{figure*}

In this section we discuss the \emph{third golden rule} of the security-by-design cycle for pattern classifiers, \ie, how to react to \emph{past} attacks and prevent \emph{future} ones. We categorize the corresponding defenses as depicted in Fig.~\ref{fig:defenses}.

\subsection{Reactive Defenses}

Reactive defenses aim to counter \emph{past} attacks. In some applications, reactive strategies may be even more convenient and effective than pure proactive approaches aimed to solely mitigate the risk of potential future attacks~\cite{barth12,biggio14-svm-chapter,biggio14-ijprai}.
Reactive approaches include: ($i$) timely detection of novel attacks, ($ii$) frequent classifier retraining, and ($iii$) verification of consistency of classifier decisions against training data and ground-truth labels~\cite{joseph13-dagstuhl,biggio14-ijprai}. In practice, to timely identify and block novel security threats, one can leverage collaborative approaches and honeypots, \ie, online services purposely vulnerable with the specific goal of collecting novel spam and malware samples.
To correctly detect recently-reported attacks, the classifier should be frequently retrained on newly-collected data (including them), and novel features and attack detectors may also be considered (see, \eg, the spam arms race discussed in Sect.~\ref{sect:arms-race}).
This procedure should also be automated to some extent to act more readily when necessary; \eg, using automatic \emph{drift} detection techniques~\cite{kuncheva08,biggio14-ijprai,jordaney17-usenix}.
The correctness of classifier decisions should finally be verified by expert domains. This raises the issue of how to involve \emph{humans in the loop} in a more coordinated manner, to supervise and verify the correct functionality of learning systems.

\subsection{Proactive Defenses} \label{sect:proactive-defenses}

Proactive defenses aim to prevent \emph{future} attacks. The main ones proposed thus far can be categorized according to the paradigms of  \emph{security by design} and \emph{security by obscurity}, as discussed in the following.

\subsubsection{Security-by-Design Defenses against White-box Attacks} \label{sect:static-defenses}

The paradigm of security by design advocates that a system should be designed from the ground up to be secure. Based on this idea, several learning algorithms have been adapted to explicitly take into account different kinds of adversarial data manipulation.
These defenses are designed in a \emph{white-box} setting in which the attacker is assumed to have perfect knowledge of the attacked system. There is thus no need to probe the targeted classifier to improve knowledge about its behavior (as instead done in gray-box and black-box attacks).

\begin{figure*}[t]
\centering
\includegraphics[width=0.21\textwidth]{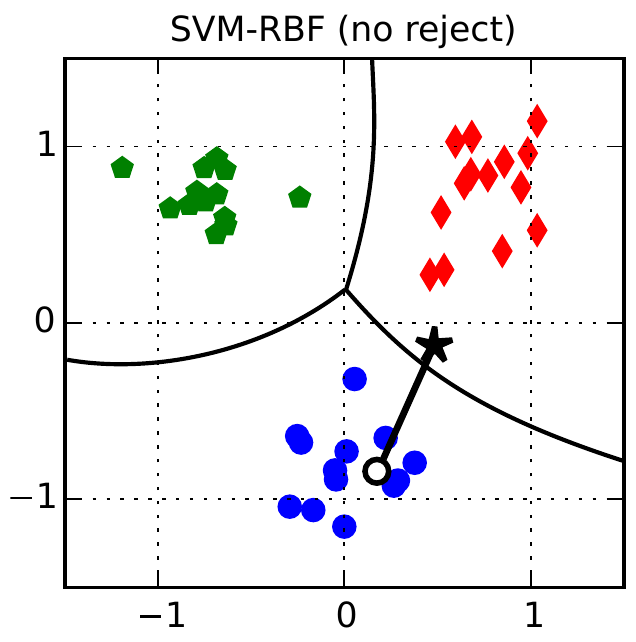}
\includegraphics[width=0.21\textwidth]{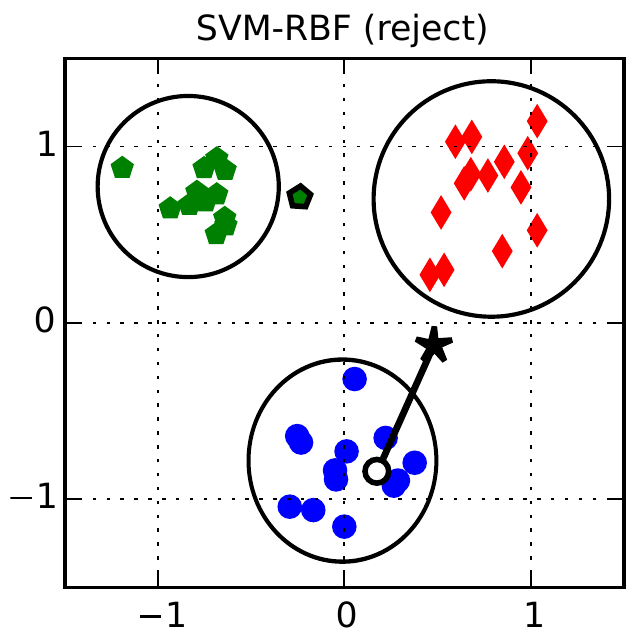}
\includegraphics[width=0.21\textwidth]{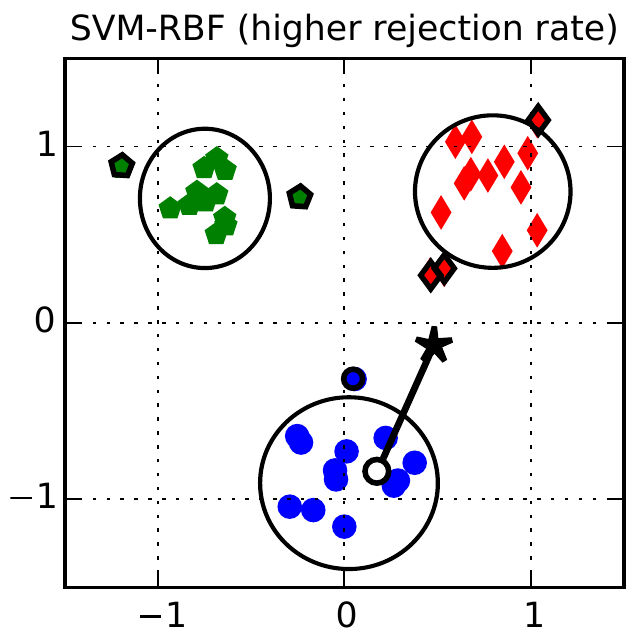}
\caption{Effect of \emph{class-enclosing} defenses against blind-spot adversarial examples on multiclass SVMs with RBF kernels, adapted from~\cite{melis17-vipar}. Rejected samples are highlighted with black contours. The adversarial example (black star) is misclassified only by the standard SVM (left plot), while SVM with rejection correctly identifies it as an adversarial example (middle plot). Rejection thresholds can be modified to increase classifier security by tightening class enclosure (right plot), at the expense of misclassifying more legitimate samples.}
\label{fig:svm-rbf-reject}
\end{figure*}

\myparagraph{Countering Evasion Attacks} In 2004, Dalvi~\etal~\cite{dalvi04} proposed the first adversary-aware classifier against evasion attacks, based on iteratively retraining the classifier on the simulated attacks. This is not very different from the idea of \emph{adversarial training} that has been recently used in deep networks to counter adversarial examples~\cite{szegedy14-iclr,goodfellow15-iclr}, or to harden decision trees and random forests~\cite{kantchelian16-icml}.
These defenses are however heuristic, with no formal guarantees on convergence and robustness properties.
More theoretically-sound approaches relying on \emph{game theory} have been proposed to overcome these limitations. Zero-sum games have been formulated to learn \emph{invariant} transformations like feature insertion, deletion and rescaling~\cite{globerson06-icml,teo08,dekel10}.\footnote{Note that similar ideas have been exploited also to model uncertainty on some parameters of the data distribution and learn optimal robust classifiers against worst-case changes of such parameters~\cite{dougherty05}.}
More rigorous approaches have then introduced Nash and Stackelberg games for secure learning, deriving formal conditions for existence and uniqueness of the game equilibrium, under the assumption that each player knows everything about the opponents and the game~\cite{liu10a,bruckner12}. Randomized players~\cite{bulo17-tnnls} and uncertainty on the players' strategies~\cite{grosshans13} have also been considered to simulate less pessimistic scenarios. 
Despite these approaches seem promising, understanding the extent to which the resulting attack strategies are representative of practical scenarios remains an open issue~\cite{wooldridge12,cybenko12}. 
Adversarial learning is not a (board) game with well-defined rules and, thus, the objective functions of real-world attackers may not even correspond to those hypothesized in the aforementioned games. It may be thus interesting to verify, reactively, whether real-world attackers behave as hypothesized, and exploit feedback from the observed attacks to improve the definition of the attack strategy. 
Another relevant problem of these approaches is their scalability to large datasets and high-dimensional feature spaces, as it may be too computationally costly to generate a sufficient number of attack samples to correctly represent their distribution, \ie, to effectively tackle the curse of dimensionality.

A more efficient approach relies on \emph{robust optimization}. Robust optimization formulates adversarial learning as a minimax problem 
in which the inner problem maximizes the training loss by manipulating the training points under worst-case, bounded perturbations, while the outer problem trains the learning algorithm to minimize the corresponding worst-case training loss~\cite{xu09,qi13,goodfellow15-iclr,madry18-iclr}.
An interesting result by Xu~\etal~\cite{xu09} has shown that  the inner problem can be solved in closed form, at least for linear SVMs, yielding a standard regularized loss formulation that penalizes the classifier parameters using the \emph{dual norm} of the input noise.
This means that different regularizers amount to hypothesizing different kinds of bounded worst-case noise on the input data.
This has effectively established an equivalence between regularized learning problems and robust optimization, which has in turn enabled approximating computationally-demanding secure learning models (\eg, game-theoretical ones) with more efficient ones based on regularizing the objective function in a specific manner~\cite{demontis16-spr,russu16-aisec,demontis17-tdsc}, also in structured learning~\cite{torkamani14-icml}.
Notably, recent work has also derived an extension of this formulation to nonlinear classifiers~\cite{kolter18-icml}.
The main effect of the aforementioned techniques is to smooth out the decision function of the classifier, making it less sensitive to worst-case input changes.
This in turn means reducing the norm of the input gradients.
More direct (and sometimes equivalent) approaches obtain the same effect by penalizing the input gradients using specific regularization terms~\cite{lyu15-icdm,sokolic17,simon18,madry18-iclr}.

Another line of defenses against evasion attacks is based on detecting and rejecting samples which are sufficiently far from the training data in feature space (similarly to the defense discussed in Sect.~\ref{sect:evasion-app})~\cite{biggio15-mcs,bendale16-cvpr,melis17-vipar,meng17-ccs,jordaney17-usenix}. These samples are usually referred to as \emph{blind-spot} evasion points, as they appear in regions of the feature space scarcely populated by training data. These regions can be assigned to any class during classifier training without any substantial increase in the training loss. In practice, this is a simple consequence of the \emph{stationarity} assumption underlying many machine-learning algorithms (according to which training and test data come from the same distribution)~\cite{moreno-torres12,pillai13-pr}, and such rejection-based defenses simply aim to overcome this issue.

Finally, we point out that \textit{classifier ensembles} have been also exploited to improve security against evasion attempts (\eg, by implementing rejection-based mechanisms or secure fusion rules)~\cite{kolcz09,biggio10-ijmlc,biggio15-mcs,corona17-esorics,wild16-pr,biggio17-tpami} and even against poisoning attacks~\cite{biggio11-mcs}. They may however worsen  security if the base classifiers are not properly \emph{combined}~\cite{biggio15-mcs,corona17-esorics}.
Many other heuristic defense techniques have also been proposed; \eg, training neural networks with bounded activation functions and training data augmentation to improve stability to input changes~\cite{zantedeschi17-aisec,zheng16-cvpr}.

\begin{figure*}[t]
	\centering
	\includegraphics[trim=20 10 20 5, clip, height=0.21\textwidth]{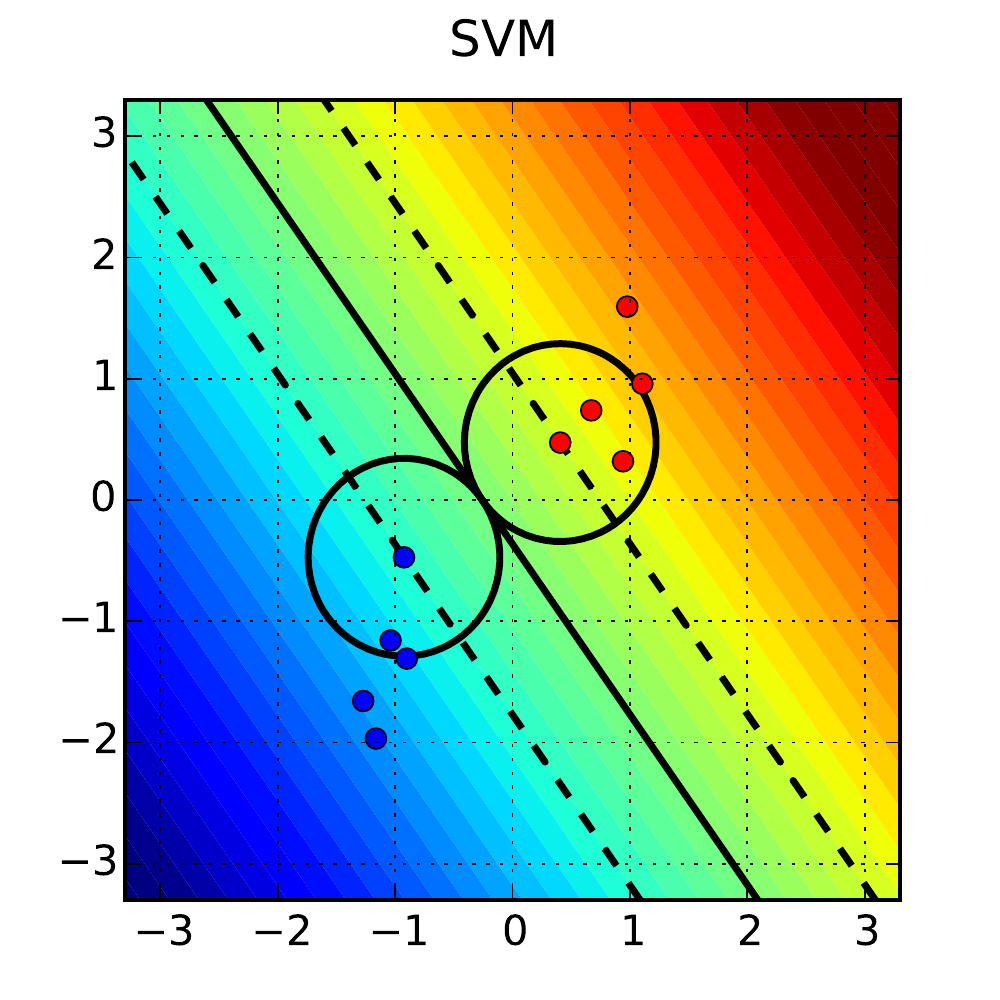}
	\includegraphics[trim=20 10 20 5, clip, height=0.21\textwidth]{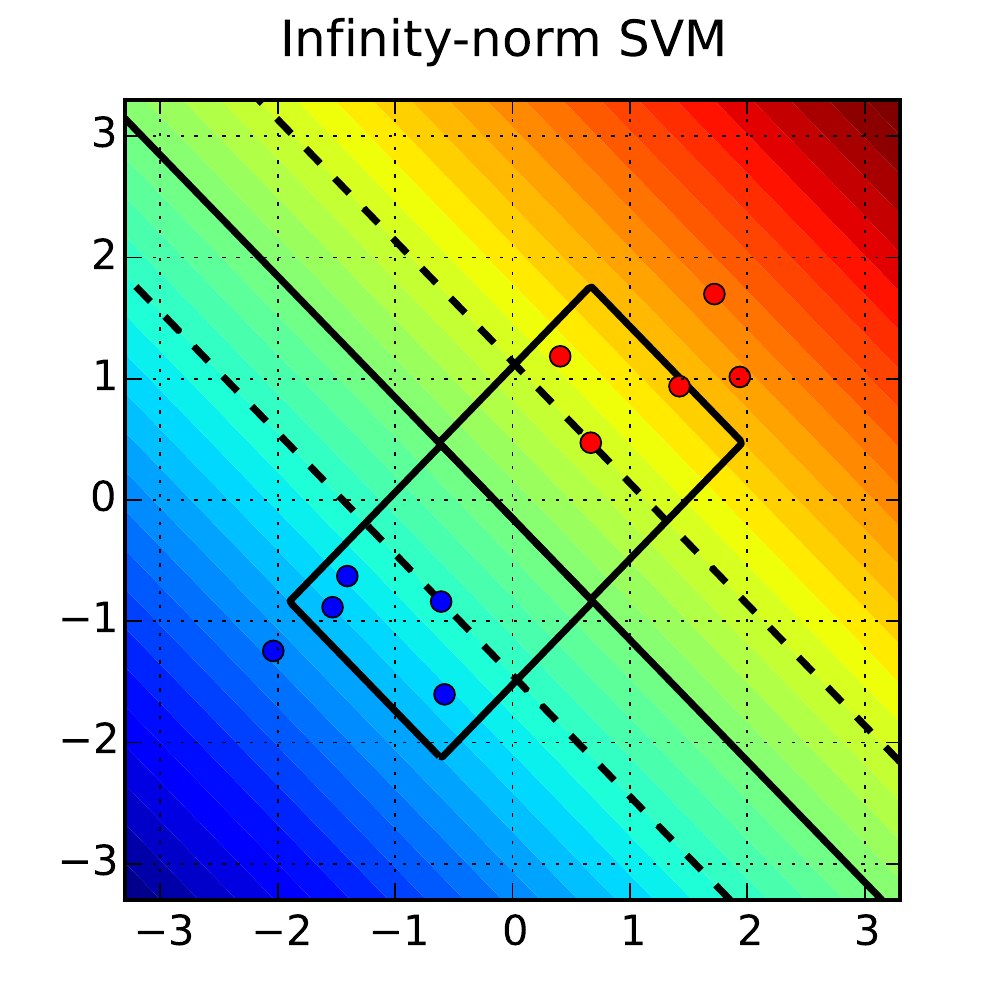}		
	\includegraphics[trim=20 10 20 5, clip, height=0.21\textwidth]{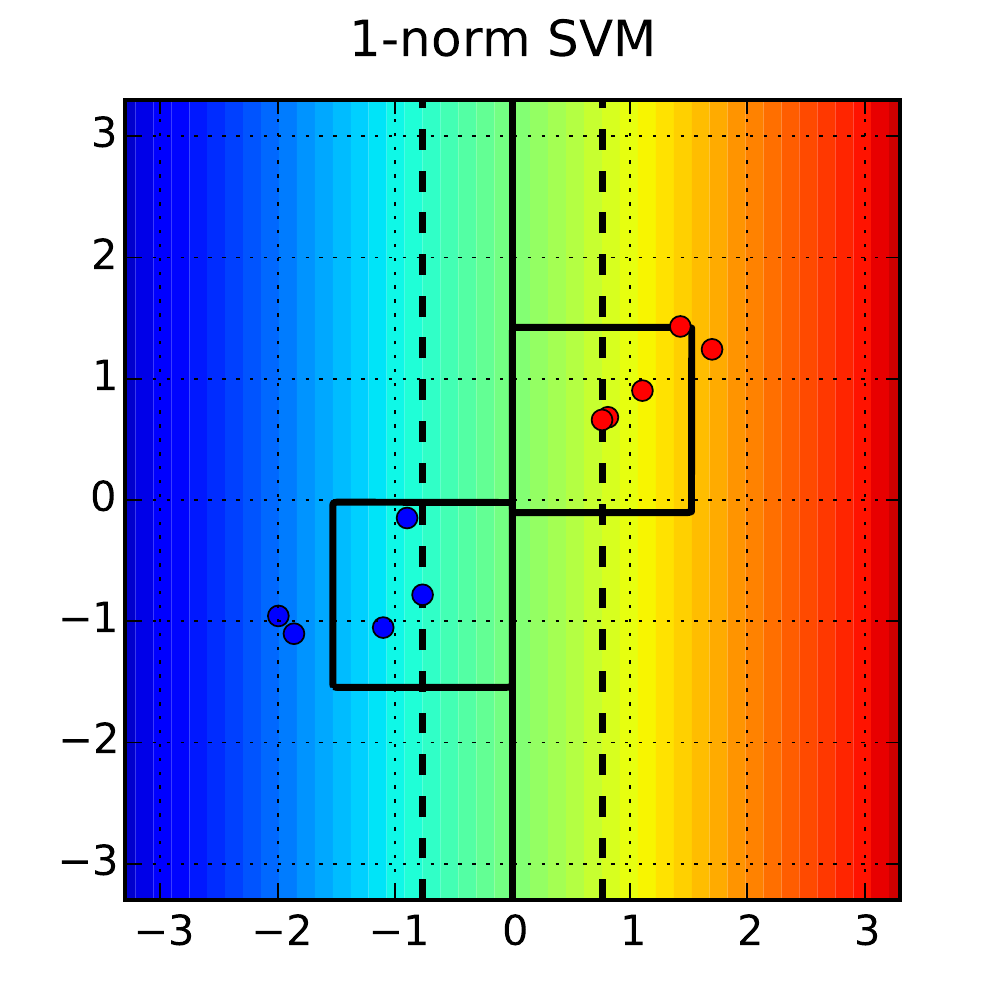}
	\includegraphics[trim=5 5 5 7, clip, height=0.21\textwidth]{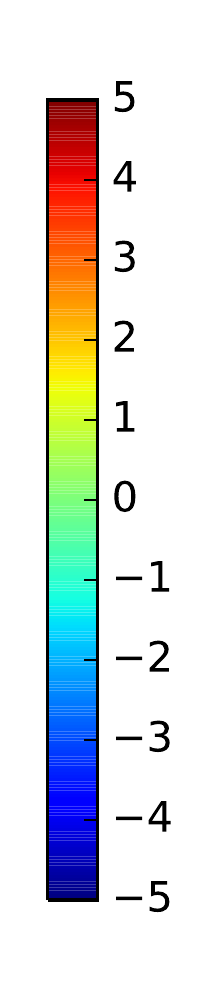}
	\caption{Decision functions for linear SVMs with $\ell_2$, $\ell_\infty$ and $\ell_1$ regularization on the feature weights~\cite{demontis16-spr,russu16-aisec}. The feasible domain of adversarial modifications (characterizing the equivalent robust optimization problem) is shown for some training points, respectively with $\ell_2$, $\ell_1$ and $\ell_\infty$ balls. Note how the shape of these balls influences the orientation of the decision boundaries, \ie, how different regularizers optimally counter specific kinds of adversarial noise.}
\label{fig:dual-norm}
\end{figure*}

\myparagraph{Effect on Decision Boundaries} We aim to discuss here how the proposed defenses substantially \emph{change} the way classifiers learn their decision boundaries.
Notably, defenses involving retraining on the attack samples and rejection mechanisms achieve security against \emph{evasion} by essentially countering blind-spot attacks. One potential effect of this assumption is that the resulting decision functions may tend to \emph{enclose} the (stationary) training classes more tightly. This in turn may require one to trade-off between the security against potential attacks and the number of misclassified  (stationary) samples at test time, as empirically shown in Sect.~\ref{sect:evasion-app}, and conceptually depicted in Fig.~\ref{fig:svm-rbf-reject}~\cite{melis17-vipar}. 
The other relevant effect, especially induced by regularization methods inspired from robust optimization, is to provide a \emph{noise-specific margin} between classes, as conceptually represented in Fig.~\ref{fig:dual-norm}~\cite{demontis16-spr,russu16-aisec}. 
These are the two main effects induced by the aforementioned secure learning approaches in feature space.

It is finally worth remarking that, by using a \emph{secure learning} algorithm, one can counter blind-spot evasion samples, but definitely not adversarial examples whose feature vectors become \emph{indistinguishable} from those of training samples belonging to different classes. 
In this case, indeed, any learning algorithm would not be able to tell such samples apart~\cite{maiorca18-arxiv}. The security properties of learning algorithms should be thus considered independently from those exhibited by the chosen feature representation. Security of features should be considered as an additional, important requirement; features should not only be discriminant, but also \emph{robust} to manipulation, to avoid straightforward classifier evasion by mimicking the feature values exhibited by legitimate samples.
In the case of deep convolutional networks, most of the problems arise from the fact that the learned mapping from input to deep space (\ie, the feature representation) violates the smoothness assumption of learning algorithms: samples that are close in input space may be very far in deep space.
In fact, as also reported in Sect.~\ref{sect:evasion-app}, adversarial examples in \emph{deep space} become indistinguishable from training samples of other classes  for sufficiently-high adversarial input perturbations~\cite{melis17-vipar}. Therefore, this vulnerability can only be patched by retraining or re-engineering the deeper layers of the network (and not only the last ones)~\cite{melis17-vipar,szegedy14-iclr}.

\myparagraph{Countering Poisoning Attacks} While most work focused on countering evasion attacks at test time, some white-box defenses have also been proposed against poisoning attacks~\cite{nelson11-aisec,nelson08,cretu08,rubinstein09,biggio11-mcs,liu17-aisec,steinhardt17-nips,jagielski18-sp}. 
To compromise a learning algorithm during training, an attack has to be exhibit different characteristics from those shown by the rest of the training data (otherwise it would have no impact at all)~\cite{biggio11-mcs}. Poisoning attacks can be thus regarded as \emph{outliers}, and countered using data sanitization (\ie, attack detection and removal)~\cite{cretu08,biggio11-mcs,steinhardt17-nips}, and robust learning (\ie, learning algorithms based on \emph{robust statistics} that are intrinsically less sensitive to outlying training samples, \eg, via bounded losses or kernel functions)~\cite{rubinstein09,liu17-aisec,xu17-pr,christmann04,jagielski18-sp,boot14-pr}.

\subsubsection{Security-by-Obscurity Defenses against Black-box Attacks}

These proactive defenses, also known as \emph{disinformation} techniques in~\cite{barreno06-asiaccs,barreno10,huang11}, follow the paradigm of \emph{security by obscurity}, \ie, they hide information to the attacker to improve security. These defenses aim to counter gray-box and black-box attacks in which probing mechanisms are used to improve surrogate models or refine evasion attempts by querying the targeted classifier.

Some examples include~\cite{joseph13-dagstuhl}: ($i$) randomizing collection of training data (collect at different timings, and locations); ($ii$) using difficult to reverse-engineer classifiers (\eg, classifier ensembles); ($iii$) denying access to the actual classifier or training data; and ($iv$) randomizing the classifier's output to give imperfect feedback to the attacker.
The latter approach has been firstly proposed in 2008~\cite{biggio08-spr} as an effective way to hide information about the classification function to the attacker, with recent follow-ups in~\cite{bulo17-tnnls,meng17-ccs} to counter adversarial examples. 
However, it is still an open issue to understand whether and to which extent randomization may be used to make it harder for the attacker to learn a proper surrogate model, and to implement  privacy-preserving  mechanisms~\cite{rubinstein12} against model inversion and hill-climbing attacks~\cite{biggio14-svm-chapter,fredrikson15-ccs,tramer16-usenix,adler05,galbally09,martinez11}.

Notably, security-by-obscurity defenses may not always be helpful. Gradient masking has been proposed to hide the gradient direction used to craft adversarial examples~\cite{papernot16-distill,lu17-iccv}, but it has been shown that it can be easily circumvented with surrogate learners~\cite{biggio13-ecml,carlini17-sp,papernot17-asiaccs}, exploiting the same principle behind attacking non-differentiable classifiers (discussed in Sect.~\ref{sect:evasion-attacks})~\cite{russu16-aisec}.

\vspace{-5pt}
\section{Conclusions and Future Work} \label{sect:conclusions}
\vspace{-5pt}
In this paper, we have presented a thorough overview of work related to the security of machine learning, pattern recognition, and deep neural networks, with the goal of providing a clearer historical picture along with useful guidelines on how to assess and improve their security against adversarial attacks.

We conclude this work by discussing some future research paths arising from the fact that machine learning has been originally developed for \emph{closed-world} problems where the possible ``states of nature'' and ``actions'' that a rationale agent can implement are perfectly known. Using the words of a famous speech by Donald Rumsfeld, one could argue that machine learning can deal with \emph{known unknowns}.\footnote{\url{http://archive.defense.gov/Transcripts/Transcript.aspx?TranscriptID=2636}}
Unfortunately, adversarial machine learning often deals with \emph{unknown unknowns}. When learning systems are deployed in adversarial environments in the \emph{open world}, they can misclassify (with high-confidence) never-before-seen inputs that are largely different from known training data. We know that \emph{unknown unknowns} are the real threat in many security problems (\eg, zero-day attacks in computer security). Although they can be mitigated using the proactive approach described in this work, they remain a primary open issue for adversarial machine learning, as modeling attacks relies on \emph{known unknowns}, while \emph{unknown unknowns} are unpredictable.

We are firmly convinced that new research paths should be explored to address this fundamental issue, complementary to formal verification and certified defenses~\cite{huang17-cav,steinhardt17-nips}. Machine learning algorithms should be able to detect \emph{unknown unknowns} using robust methods for anomaly or novelty detection, potentially asking for human intervention when required. The development of practical methods for explaining, visualizing and interpreting the operation of machine-learning systems could also help system designers to investigate the behavior of such systems on cases that are not statistically represented by the training data, and decide whether to trust their decisions on such \emph{unknown unknowns} or not. 
These future research paths lie at the intersection of the field of adversarial machine learning and the emerging fields of robust artificial intelligence and interpretability of machine learning~\cite{dietterich17,lipton16}. We believe that these directions will help our society to get a more conscious understanding of the potential and limits of modern data-driven machine-learning technologies. 

\vspace{-5pt}
\section*{Acknowledgments}
\vspace{-5pt}
We are grateful to Ambra Demontis and Marco Melis for providing the experimental results on evasion and poisoning attacks, and to Ian Goodfellow for providing feedback on the timeline of evasion attacks. This work was also partly supported by the EU H2020 project ALOHA, under the European Union's Horizon 2020 research and innovation programme (grant no. 780788).


\begin{thebibliography}{10}
\expandafter\ifx\csname url\endcsname\relax
  \def\url#1{\texttt{#1}}\fi
\expandafter\ifx\csname urlprefix\endcsname\relax\def\urlprefix{URL }\fi
\expandafter\ifx\csname href\endcsname\relax
  \def\href#1#2{#2} \def\path#1{#1}\fi

\bibitem{gu18-pr}
J.~Gu, Z.~Wang, J.~Kuen, L.~Ma, A.~Shahroudy, B.~Shuai, T.~Liu, X.~Wang,
  G.~Wang, J.~Cai, T.~Chen, Recent advances in convolutional neural networks,
  Pattern Recognition 77 (2018) 354 -- 377.
    
\bibitem{szegedy14-iclr}
C.~Szegedy, W.~Zaremba, I.~Sutskever, J.~Bruna, D.~Erhan, I.~Goodfellow,
  R.~Fergus, \href{http://arxiv.org/abs/1312.6199}{Intriguing properties of
  neural networks}, in: ICLR, 2014.

\bibitem{goodfellow15-iclr}
I.~J. Goodfellow, J.~Shlens, C.~Szegedy, Explaining and harnessing adversarial
  examples, in: ICLR, 2015.

\bibitem{nguyen15-cvpr}
A.~M. Nguyen, J.~Yosinski, J.~Clune, Deep neural networks are easily fooled:
  High confidence predictions for unrecognizable images., in: IEEE CVPR, 2015, pp. 427--436.

\bibitem{moosavi16-deepfool}
S.-M. Moosavi-Dezfooli, A.~Fawzi, P.~Frossard, Deepfool: a simple and accurate
  method to fool deep neural networks, in: IEEE CVPR, 2016, pp. 2574--2582.

\bibitem{papernot16-distill}
N.~Papernot, P.~McDaniel, X.~Wu, S.~Jha, A.~Swami, Distillation as a defense to
  adversarial perturbations against deep neural networks, in: IEEE
  Symp. Security \& Privacy (SP), 2016, pp. 582--597.

\bibitem{melis17-vipar}
M.~Melis, A.~Demontis, B.~Biggio, G.~Brown, G.~Fumera, F.~Roli, Is deep
  learning safe for robot vision? Adversarial examples against the iCub
  humanoid, in: ICCV Workshop {ViPAR}, 2017.

\bibitem{meng17-ccs}
D.~Meng, H.~Chen, {MagNet}: a two-pronged defense against adversarial examples,
  in: 24th ACM Conf. Computer and Comm. Sec. (CCS), 2017.

\bibitem{lu17-iccv}
J.~Lu, T.~Issaranon, D.~Forsyth, Safetynet: Detecting and rejecting adversarial
  examples robustly, in: IEEE ICCV, 2017.

\bibitem{li17-iccv}
X.~Li, F.~Li, Adversarial examples detection in deep networks with
  convolutional filter statistics, in: IEEE ICCV, 2017.

\bibitem{grosse17-esorics}
K.~Grosse, N.~Papernot, P.~Manoharan, M.~Backes, P.~D. McDaniel, Adversarial
  examples for malware detection, in: {ESORICS} {(2)}, Vol. 10493 of LNCS,
  Springer, 2017, pp. 62--79.

\bibitem{jordaney17-usenix}
R.~Jordaney, K.~Sharad, S.~K.~Dash, Z.~Wang, D. Papini, I.~Nouretdinov, L.~Cavallaro, Transcend: Detecting concept drift in malware classification models, USENIX Sec. Symp.,
  USENIX Assoc., 2017, pp 625?642.

\bibitem{mcdaniel16}
P.~McDaniel, N.~Papernot, Z.~B. Celik, Machine learning in adversarial
  settings, IEEE Security {\&} Privacy 14~(3) (2016) 68--72.

\bibitem{papernot16-sp}
N.~Papernot, P.~McDaniel, S.~Jha, M.~Fredrikson, Z.~B. Celik, A.~Swami, The
  limitations of deep learning in adversarial settings, in: 1st IEEE
  European Symp. Security and Privacy, 2016, pp. 372--387.

\bibitem{papernot17-asiaccs}
N.~Papernot, P.~McDaniel, I.~Goodfellow, S.~Jha, Z.~B. Celik, A.~Swami,
  Practical black-box attacks against machine learning, in: ASIA CCS '17, ACM,  2017, pp. 506--519.

\bibitem{carlini17-sp}
N.~Carlini, D.~A. Wagner, Towards evaluating the robustness of neural networks,
  in: IEEE  Symp. Security \& Privacy (SP), 2017, pp. 39--57.

\bibitem{sharif16-ccs}
M.~Sharif, S.~Bhagavatula, L.~Bauer, M.~K. Reiter, Accessorize to a crime: Real
  and stealthy attacks on state-of-the-art face recognition, in: Conf. Computer and Comm. Security (CCS), ACM,
  2016, pp. 1528--1540.

\bibitem{xie17-iccv}
C.~Xie, J.~Wang, Z.~Zhang, Y.~Zhou, L.~Xie, A.~Yuille, Adversarial examples for
  semantic segmentation and object detection, in: IEEE ICCV, 2017.

\bibitem{dalvi04}
N.~Dalvi, P.~Domingos, Mausam, S.~Sanghai, D.~Verma, Adversarial
  classification, in: Int'l Conf. Knowl. Disc. and Data Mining, 2004, pp. 99--108.

\bibitem{lowd05}
D.~Lowd, C.~Meek, Adversarial learning, in:  Int'l Conf. Knowl. Disc. and Data Mining, ACM Press, Chicago,
  IL, USA, 2005, pp. 641--647.

\bibitem{lowd05-ceas}
D.~Lowd, C.~Meek, Good word attacks on statistical spam filters, in: 2nd
  Conf. Email and Anti-Spam (CEAS), Mountain View, CA, USA, 2005.

\bibitem{matsumoto02}
T.~Matsumoto, H.~Matsumoto, K.~Yamada, S.~Hoshino, Impact of artificial ``gummy''
  fingers on fingerprint systems, Datensch. und Datensich. 26~(8).
  
\bibitem{barreno06-asiaccs}
M.~Barreno, B.~Nelson, R.~Sears, A.~D. Joseph, J.~D. Tygar, Can machine
  learning be secure?, in: ASIA CCS '06, ACM, 2006, pp. 16--25.

\bibitem{nelson08}
B.~Nelson, M.~Barreno, F.~J. Chi, A.~D. Joseph, B.~I.~P. Rubinstein, U.~Saini,
  C.~Sutton, J.~D. Tygar, K.~Xia, Exploiting machine learning to subvert your
  spam filter, in: LEET '08, USENIX Assoc., 2008, pp. 1--9.

\bibitem{rubinstein09}
B.~I. Rubinstein, B.~Nelson, L.~Huang, A.~D. Joseph, S.-h. Lau, S.~Rao,
  N.~Taft, J.~D. Tygar, Antidote: understanding and defending against poisoning
  of anomaly detectors, in: IMC '09, ACM, 2009, pp. 1--14.

\bibitem{biggio12-icml}
B.~Biggio, B.~Nelson, P.~Laskov, Poisoning attacks against support vector
  machines, in: 29th ICML, 2012, pp. 1807--1814.

\bibitem{kloft10}
M.~Kloft, P.~Laskov, Online anomaly detection under adversarial impact, in:
13th AISTATS, 2010, pp. 405--412.

\bibitem{kloft12b}
M.~Kloft, P.~Laskov, Security analysis of online centroid anomaly detection,
  JMLR 13 (2012) 3647--3690.

\bibitem{biggio15-icml}
H.~Xiao, B.~Biggio, G.~Brown, G.~Fumera, C.~Eckert, F.~Roli, Is feature
  selection secure against training data poisoning?, in: 32nd ICML, Vol.~37,
  2015, pp. 1689--1698.

\bibitem{biggio14-svm-chapter}
B.~Biggio, I.~Corona, B.~Nelson, B.~Rubinstein, D.~Maiorca, G.~Fumera,
  G.~Giacinto, F.~Roli,
  Security evaluation of
  support vector machines in adversarial environments, in: Y.~Ma, G.~Guo
  (Eds.), Support Vector Machines Applications, Springer Int'l
  Publishing, Cham, 2014, pp. 105--153.


\bibitem{mei15-aaai}
S.~Mei, X.~Zhu, Using machine teaching to identify optimal training-set attacks
  on machine learners, in: 29th AAAI, 2015.

\bibitem{koh17-icml}
P.~W. Koh, P.~Liang, Understanding black-box predictions via influence
  functions, in: ICML, 2017.

\bibitem{biggio17-aisec}
L.~{Mu\~noz-Gonz\'alez}, B.~Biggio, A.~Demontis, A.~Paudice, V.~Wongrassamee,
  E.~C. Lupu, F.~Roli, Towards poisoning of deep learning algorithms with
  back-gradient optimization, in: AISec '17, ACM, 2018, pp.
  27--38.

\bibitem{wittel04}
G.~L. Wittel, S.~F. Wu, On attacking statistical spam filters, in: 1st
  Conf. Email and Anti-Spam (CEAS), 2004.

\bibitem{globerson06-icml}
A.~Globerson, S.~T. Roweis, Nightmare at test time: robust learning by feature
  deletion, in: 23rd  ICML, Vol. 148, ACM, 2006, pp. 353--360.

\bibitem{teo08}
C.~H. Teo, A.~Globerson, S.~Roweis, A.~Smola, Convex learning with invariances,
  in: NIPS 20, MIT Press, 2008, pp. 1489--1496.

\bibitem{dekel10}
O.~Dekel, O.~Shamir, L.~Xiao,
  \href{http://dx.doi.org/10.1007/s10994-009-5124-8}{Learning to classify with
  missing and corrupted features}, Machine Learning 81 (2010) 149--178.

\bibitem{biggio13-ecml}
B.~Biggio, I.~Corona, D.~Maiorca, B.~Nelson, N.~\v{S}rndi\'{c}, P.~Laskov,
  G.~Giacinto, F.~Roli, Evasion attacks against machine learning at test time,
  in: ECML PKDD, Part III, Vol. 8190 of LNCS, Springer, 2013, pp. 387--402.

\bibitem{srndic14}
N.~\v{S}rndic, P.~Laskov, Practical evasion of a learning-based classifier: A
  case study, in: IEEE Symp. Security and Privacy, SP '14, 2014, pp. 197--211.

\bibitem{barreno10}
M.~Barreno, B.~Nelson, A.~Joseph, J.~Tygar, The security of machine learning,
  Machine Learning 81 (2010) 121--148.

\bibitem{biggio14-tkde}
B.~Biggio, G.~Fumera, F.~Roli, Security evaluation of pattern classifiers under
  attack, IEEE Trans. Knowl. and Data Eng. 26~(4) (2014)
  984--996.

\bibitem{biggio14-ijprai}
B.~Biggio, G.~Fumera, F.~Roli, Pattern recognition systems under attack: Design
  issues and research challenges, IJPRAI 28~(7)
  (2014) 1460002.

\bibitem{biggio15-spmag}
B.~Biggio, G.~Fumera, P.~Russu, L.~Didaci, F.~Roli, Adversarial biometric
  recognition : A review on biometric system security from the adversarial
  machine-learning perspective, IEEE Signal Proc. Mag., 32~(5) (2015)
  31--41.

\bibitem{kolcz09}
A.~Kolcz, C.~H. Teo, Feature weighting for improved classifier robustness, in:
  6th Conf. Email and Anti-Spam (CEAS), 2009.

\bibitem{biggio08-spr}
B.~Biggio, G.~Fumera, F.~Roli, Adversarial pattern classification using
  multiple classifiers and randomisation, in: SSPR 2008, Vol.
  5342 of LNCS, Springer, 2008, pp. 500--509.

\bibitem{bruckner12}
M.~Br\"{u}ckner, C.~Kanzow, T.~Scheffer, Static prediction games for
  adversarial learning problems, JMLR 13 (2012) 2617--2654.

\bibitem{bulo17-tnnls}
S.~{Rota Bul\`o}, B.~Biggio, I.~Pillai, M.~Pelillo, F.~Roli, Randomized
  prediction games for adversarial machine learning, IEEE Trans.
  Neural Networks and Learning Systems 28~(11) (2017) 2466--2478.

\bibitem{demontis17-tdsc}
A.~Demontis, M.~Melis, B.~Biggio, D.~Maiorca, D.~Arp, K.~Rieck, I.~Corona,
  G.~Giacinto, F.~Roli, Yes, machine learning can be more secure! A case study
  on android malware detection, IEEE Trans. Dep. and Secure Comp.

\bibitem{nips07-adv}
P.~Laskov, R.~Lippmann (Eds.),
  \href{http://mls-nips07.first.fraunhofer.de}{{NIPS} {W}orkshop on {M}achine
  {L}earning in {A}dversarial {E}nvironments for {C}omputer {S}ecurity}, 2007.

\bibitem{laskov10-ed}
P.~Laskov, R.~Lippmann,
  Machine learning in
  adversarial environments, Machine Learning 81 (2010) 115--119.

\bibitem{joseph13-dagstuhl}
A.~D. Joseph, P.~Laskov, F.~Roli, J.~D. Tygar, B.~Nelson, {Machine Learning
  Methods for Computer Security (Dagstuhl Perspectives Workshop 12371)},
  Dagstuhl Manifestos 3~(1) (2013) 1--30.

\bibitem{biggio17-aisec-proc}
B.~M. Thuraisingham, B.~Biggio, D.~M. Freeman, B.~Miller, A.~Sinha (Eds.),
  AISec '17: 10th Workshop on AI and Security, ACM, 2017.

\bibitem{joseph18-advml-book}
A.~D. Joseph, B.~Nelson, B.~I.~P. Rubinstein, J.~Tygar, Adversarial Machine
  Learning, Cambridge University Press, 2018.

\bibitem{corona13-is}
I.~Corona, G.~Giacinto, F.~Roli, Adversarial attacks against intrusion
  detection systems: Taxonomy, solutions and open issues, Information Sciences
  239~(0) (2013) 201 -- 225.

\bibitem{han16-ccs}
X.~Han, N.~Kheir, D.~Balzarotti, Phisheye: Live monitoring of sandboxed
  phishing kits, in: ACM CCS, 2016, pp. 1402--1413.

\bibitem{corona17-esorics}
I.~Corona, B.~Biggio, M.~Contini, L.~Piras, R.~Corda, M.~Mereu, G.~Mureddu,
  D.~Ariu, F.~Roli, Deltaphish: Detecting phishing webpages in compromised
  websites, in: ESORICS, Vol. 10492 of LNCS,
  Springer, 2017, pp. 370--388.

\bibitem{biggio11-prl}
B.~Biggio, G.~Fumera, I.~Pillai, F.~Roli,
 A survey and experimental evaluation of image spam filtering techniques,
  PRL 32~(10) (2011) 1436 -- 1446.

\bibitem{attar13}
A.~Attar, R.~M. Rad, R.~E. Atani,
  A survey of image spamming
  and filtering techniques, Artif. Intell. Rev. 40~(1) (2013) 71--105.

\bibitem{fumera06}
G.~Fumera, I.~Pillai, F.~Roli, Spam filtering based on the analysis of text
  information embedded into images, JMLR 7 (2006) 2699--2720.

\bibitem{thomas09}
A.~O. Thomas, A.~Rusu, V.~Govindaraju, Synthetic handwritten captchas, Pattern
  Recognition 42~(12) (2009) 3365 -- 3373, new Frontiers in Handwriting
  Recognition.

\bibitem{huang11}
L.~Huang, A.~D. Joseph, B.~Nelson, B.~Rubinstein, J.~D. Tygar, Adversarial
  machine learning, in: 4th AISec, Chicago, IL, USA, 2011, pp. 43--57.

\bibitem{biggio13-aisec}
B.~Biggio, I.~Pillai, S.~R. Bul\`o, D.~Ariu, M.~Pelillo, F.~Roli, Is data
  clustering in adversarial settings secure?, in: AISec '13, ACM, 2013, pp. 87--98.

\bibitem{biggio14-aisec}
B.~Biggio, K.~Rieck, D.~Ariu, C.~Wressnegger, I.~Corona, G.~Giacinto, F.~Roli,
  Poisoning behavioral malware clustering, in: AISec '14, {ACM}, 2014, pp.
  27--36.

\bibitem{biggio14-spr}
B.~Biggio, S.~R. Bul\`o, I.~Pillai, M.~Mura, E.~Z. Mequanint, M.~Pelillo,
  F.~Roli, Poisoning complete-linkage hierarchical clustering, in: SSPR, Vol. 8621 of
  LNCS, Springer, 2014, pp. 42--52.

\bibitem{zhang16-tcyb}
F.~Zhang, P.~Chan, B.~Biggio, D.~Yeung, F.~Roli, Adversarial feature selection
  against evasion attacks, IEEE Trans. Cyb. 46~(3) (2016)
  766--777.

\bibitem{tramer16-usenix}
F.~Tram{\`e}r, F.~Zhang, A.~Juels, M.~K. Reiter, T.~Ristenpart, Stealing
  machine learning models via prediction APIs, in: USENIX Sec. Symp.,
  USENIX Assoc., 2016, pp. 601--618.

\bibitem{xu16-ndss}
W.~Xu, Y.~Qi, D.~Evans, Automatically evading classifiers, in: 
  Annual Network \& Distr. Sys. Sec. Symp. ({NDSS}), The
  Internet Society, 2016.

\bibitem{chen17-aisec}
P.-Y. Chen, H.~Zhang, Y.~Sharma, J.~Yi, C.-J. Hsieh, Zoo: Zeroth order
  optimization based black-box attacks to deep neural networks without training
  substitute models, in: AISec '17, ACM, 2017, pp. 15--26.

\bibitem{dang17-ccs}
H.~Dang, Y.~Huang, E.~Chang, Evading classifiers by morphing in the dark, in:
  ACM CCS '17, {ACM}, 2017, pp. 119--133.

\bibitem{nelson12-jmlr}
B.~Nelson, B.~I. Rubinstein, L.~Huang, A.~D. Joseph, S.~J. Lee, S.~Rao, J.~D.
  Tygar, Query strategies for evading convex-inducing classifiers, JMLR 13 (2012) 1293--1332.

\bibitem{gu17}
T.~Gu, B.~Dolan{-}Gavitt, S.~Garg, Badnets: Identifying vulnerabilities in the
  machine learning model supply chain, in: NIPS Workshop Mach. Learn. Comp. Sec., Vol. abs/1708.06733, 2017.

\bibitem{chen17}
X.~{Chen}, C.~{Liu}, B.~{Li}, K.~{Lu}, D.~{Song}, Targeted backdoor attacks on
  deep learning systems using data poisoning, ArXiv e-prints abs/1712.05526.
  
\bibitem{fredrikson15-ccs}
M.~Fredrikson, S.~Jha, T.~Ristenpart, Model inversion attacks that exploit
  confidence information and basic countermeasures, in: ACM CCS '15, ACM,
  2015, pp. 1322--1333.

\bibitem{adler05}
A.~Adler, Vulnerabilities in biometric encryption systems, in: T.~Kanade, A.~K.
  Jain, N.~K. Ratha (Eds.), 5th Int'l Conf. Audio- and
  Video-Based Biometric Person Auth., Vol. 3546 of LNCS,
  Springer, 2005, pp. 1100--1109.

\bibitem{galbally09}
J.~Galbally, C.~McCool, J.~Fierrez, S.~Marcel, J.~Ortega-Garcia, On the
  vulnerability of face verification systems to hill-climbing attacks, Patt.
  Rec. 43~(3) (2010) 1027--1038.

\bibitem{martinez11}
M.~Martinez-Diaz, J.~Fierrez, J.~Galbally, J.~Ortega-Garcia,
 An
  evaluation of indirect attacks and countermeasures in fingerprint
  verification systems, Patt. Rec. Lett. 32~(12) (2011) 1643 --
  1651.

\bibitem{demontis16-spr}
A.~Demontis, P.~Russu, B.~Biggio, G.~Fumera, F.~Roli, On security and sparsity
  of linear classifiers for adversarial settings, in: SSPR, Vol. 10029 of
  LNCS, Springer, 2016, pp. 322--332.

\bibitem{russu16-aisec}
P.~Russu, A.~Demontis, B.~Biggio, G.~Fumera, F.~Roli, Secure kernel machines
  against evasion attacks, in: AISec '16, ACM, 2016, pp. 59--69.

\bibitem{kantchelian16-icml}
A.~Kantchelian, J.~D. Tygar, A.~D. Joseph, Evasion and hardening of tree
  ensemble classifiers, in: {ICML}, Vol.~48 {JMLR} W\&CP, 2016, pp. 2387--2396.

\bibitem{athalye18-iclr}
A.~Athalye, L.~Engstrom, A.~Ilyas, K.~Kwok, Synthesizing robust adversarial
  examples, in: ICLR, 2018.

\bibitem{fogla06}
P.~Fogla, M.~Sharif, R.~Perdisci, O.~Kolesnikov, W.~Lee, Polymorphic blending
  attacks, in: USENIX
  Sec. Symp., 2006, pp.
  241--256.

\bibitem{biggio10-ijmlc}
B.~Biggio, G.~Fumera, F.~Roli, Multiple classifier systems for robust
  classifier design in adversarial environments, Int'l JMLC 1~(1) (2010) 27--41.

\bibitem{srndic13-ndss}
N.~\v{S}rndi\'{c}, P.~Laskov, Detection of malicious pdf files based on
  hierarchical document structure, in: 20th NDSS, The Internet Society, 2013.

\bibitem{xu09}
H.~Xu, C.~Caramanis, S.~Mannor,
  Robustness and
  regularization of support vector machines, JMLR 10 (2009) 1485--1510.

\bibitem{carlini17-aisec}
N.~Carlini, D.~A. Wagner, Adversarial examples are not easily detected:
  Bypassing ten detection methods, in: AISec '17, {ACM}, 2017, pp.
  3--14.
  
\bibitem{athalye18}
A.~Athalye, N.~Carlini, D.~A. Wagner, Obfuscated gradients give a false sense
  of security: Circumventing defenses to adversarial examples, in: {ICML},
  Vol.~80 of {JMLR} W\&CP, JMLR.org, 2018, pp.
  274--283.

\bibitem{dong18-cvpr}
Y.~Dong, F.~Liao, T.~Pang, X.~Hu, J.~Zhu, Boosting adversarial examples with
  momentum, in: CVPR, 2018.

\bibitem{huang17-cav}
X.~Huang, M.~Kwiatkowska, S.~Wang, M.~Wu,
  Safety verification of
  deep neural networks, in: 29th Int'l Conf. Computer Aided
  Verification, Part {I}, Vol. 10426 of LNCS, Springer, 2017, pp. 3--29.

\bibitem{pei17}
K.~Pei, Y.~Cao, J.~Yang, S.~Jana, Deepxplore: Automated whitebox testing of
  deep learning systems, in: 26th SOSP, ACM, 2017, pp. 1--18.

\bibitem{newsome06}
J.~Newsome, B.~Karp, D.~Song, Paragraph: Thwarting signature learning by
  training maliciously, in: RAID, LNCS,
  Springer, 2006, pp. 81--105.

\bibitem{barth12}
A.~Barth, B.~I. Rubinstein, M.~Sundararajan, J.~C. Mitchell, D.~Song, P.~L.
  Bartlett, A learning-based approach to reactive security, IEEE Trans. Dependable and Sec. Comp. 9~(4) (2012) 482--493.

\bibitem{kuncheva08}
L.~I. Kuncheva, Classifier ensembles for detecting concept change in streaming
  data: Overview and perspectives, in: SUEMA, 2008, pp. 5--10.

\bibitem{dougherty05}
E.~R. Dougherty, J.~Hua, Z.~Xiong, Y.~Chen, Optimal robust classifiers, Pattern Recognition 38~(10) (2005) 1520--1532.
  
\bibitem{liu10a}
W.~Liu, S.~Chawla, Mining adversarial patterns via regularized loss
  minimization, Machine Learning 81~(1) (2010) 69--83.

\bibitem{grosshans13}
M.~Gro{\ss}hans, C.~Sawade, M.~Br\"{u}ckner, T.~Scheffer, Bayesian games for
  adversarial regression problems, in: 30th ICML, JMLR W\&CP, Vol.~28,
  2013, pp. 55--63.

\bibitem{wooldridge12}
M.~Wooldridge, Does game theory work?, IEEE IS 27~(6) (2012) 76--80.

\bibitem{cybenko12}
G.~Cybenko, C.~E. Landwehr, Security analytics and measurements, IEEE Security
  {\&} Privacy 10~(3) (2012) 5--8.

\bibitem{qi13}
Z.~Qi, Y.~Tian, Y.~Shi, Robust twin support vector machine for pattern
  classification, Pattern Recognition 46~(1) (2013) 305 -- 316.

\bibitem{torkamani14-icml}
M.~A. Torkamani, D.~Lowd, On robustness and regularization of structural
  support vector machines, in:  ICML, Vol.~32 of PMLR, 2014, pp. 577--585.

\bibitem{kolter18-icml}
E.~Wong, J.~Z. Kolter, Provable defenses against adversarial examples via the
  convex outer adversarial polytope, in: {ICML}, Vol.~80 of {JMLR} W\&CP, JMLR.org, 2018, pp. 5283--5292.


\bibitem{lyu15-icdm}
C.~Lyu, K.~Huang, H.-N. Liang, A unified gradient regularization family for
  adversarial examples, in: ICDM, Vol.~00, IEEE CS, 2015, pp.
  301--309.

\bibitem{sokolic17}
J.~Sokoli{\'c}, R.~Giryes, G.~Sapiro, M.~R.~D. Rodrigues, Robust large margin deep neural networks, IEEE Trans. Signal Proc. 65~(16) (2017) 4265--4280.

\bibitem{simon18}
C.~J. Simon-Gabriel, Y.~Ollivier, B.~Sch{\"o}lkopf, L.~Bottou, D.~Lopez-Paz,
  Adversarial vulnerability of neural networks increases with input dimension, ArXiv e-prints, 2018.
  
\bibitem{madry18-iclr}
A.~{Madry}, A.~{Makelov}, L.~{Schmidt}, D.~{Tsipras}, A.~{Vladu}, {Towards Deep
  Learning Models Resistant to Adversarial Attacks}, ICLR, 2018.

\bibitem{biggio15-mcs}
B.~Biggio, I.~Corona, Z.-M. He, P. Chan, G.~Giacinto, D. Yeung,
  F.~Roli, One-and-a-half-class multiple classifier systems for secure learning
  against evasion attacks at test time, in: MCS, Vol. 9132 of LNCS, Springer, 2015, pp. 168--180.

\bibitem{bendale16-cvpr}
A.~Bendale, T.~E. Boult, Towards open set deep networks, in: IEEE CVPR, 2016, pp. 1563--1572.

\bibitem{moreno-torres12}
J.~G. Moreno-Torres, T.~Raeder, R.~Alaiz-Rodri-guez, N.~V. Chawla, F.~Herrera,
  A unifying view on dataset shift in classification, Patt. Rec.
  45~(1) (2012) 521 -- 530.
 
 \bibitem{pillai13-pr}
I.~Pillai, G.~Fumera, F.~Roli, Multi-label classification with a reject option, Patt. Rec. 46~(8) (2013) 2256 -- 2266.

\bibitem{wild16-pr}
P.~Wild, P.~Radu, L.~Chen, J.~Ferryman, Robust multimodal face and fingerprint fusion in the presence of spoofing attacks, Patt. Rec. 50 (2016) 17 -- 25.

\bibitem{biggio17-tpami}
B.~Biggio, G.~Fumera, G.~L. Marcialis, F.~Roli, Statistical meta-analysis of
  presentation attacks for secure multibiometric systems, IEEE Trans. Patt. Analysis Mach. Intell. 39~(3) (2017) 561--575.
   
\bibitem{biggio11-mcs}
B.~Biggio, I.~Corona, G.~Fumera, G.~Giacinto, F.~Roli, Bagging classifiers for
  fighting poisoning attacks in adversarial classification tasks, in: MCS, Vol. 6713 of LNCS, Springer-Verlag, 2011, pp. 350--359.

\bibitem{zantedeschi17-aisec}
V.~Zantedeschi, M.~Nicolae, A.~Rawat, Efficient defenses against adversarial
  attacks, in: AISec, {ACM}, 2017, pp. 39--49.

\bibitem{zheng16-cvpr}
S.~Zheng, Y.~Song, T.~Leung, I.~Goodfellow, Improving the robustness of deep
  neural networks via stability training, in: CVPR, 2016, pp. 4480--4488.

\bibitem{maiorca18-arxiv}
D.~Maiorca, B.~Biggio, M.~E. Chiappe, G.~Giacinto, Adversarial detection of
  flash malware: Limitations and open issues, CoRR abs/1710.10225.

\bibitem{nelson11-aisec}
B.~Nelson, B.~Biggio, P.~Laskov, Understanding the risk factors of learning in
  adversarial environments, in: AISec '11, 2011, pp. 87--92.

\bibitem{cretu08}
G.~F. Cretu, A.~Stavrou, M.~E. Locasto, S.~J. Stolfo, A.~D. Keromytis, Casting
  out demons: Sanitizing training data for anomaly sensors, in: IEEE Symp. Security and Privacy, IEEE CS, 2008,
  pp. 81--95.

\bibitem{liu17-aisec}
C.~Liu, B.~Li, Y.~Vorobeychik, A.~Oprea, Robust linear regression against
  training data poisoning, in: AISec '17, {ACM}, 2017, pp. 91--102.

\bibitem{steinhardt17-nips}
J.~Steinhardt, P.~W.~Koh, P.~Liang, Certified defenses for data poisoning attacks, in: NIPS, 2017.

\bibitem{jagielski18-sp}
M.~Jagielski, A.~Oprea, B.~Biggio, C.~Liu, C.~Nita-Rotaru, B.~Li, Manipulating
  machine learning: Poisoning attacks and countermeasures for regression
  learning, in: 39th IEEE Symp. Security and Privacy, 2018.

\bibitem{xu17-pr}
G.~Xu, Z.~Cao, B.-G. Hu, J.~C. Principe, Robust support vector machines based
  on the rescaled hinge loss function, Patt. Rec. 63 (2017) 139 --
  148.

\bibitem{christmann04}
A.~Christmann, I.~Steinwart, On robust properties of convex risk minimization methods for pattern recognition, JMLR 5 (2004) 1007--1034.

\bibitem{boot14-pr}
J.~Bootkrajang, A.~Kaban, Learning kernel logistic regression in the presence of class label noise, Patt. Rec. 47~(11) (2014) 3641 -- 3655.
  
\bibitem{rubinstein12}
B.~I.~P. Rubinstein, P.~L. Bartlett, L.~Huang, N.~Taft, Learning in a large function space: Privacy-preserving mechanisms for SVM learning, J. Privacy and Conf. 4~(1) (2012) 65--100.

\bibitem{dietterich17}
T.~Dietterich, Steps Toward Robust Artificial Intelligence, AI Magazine 38~(3), 2017.

\bibitem{lipton16}
Z.~Lipton, The mythos of model interpretability, ICML Workshop on Human Interpretability of Machine Learning, 2016.
\end{thebibliography}

\par\noindent 
\parbox[t]{\linewidth}{
\noindent\parpic{\includegraphics[height=1in]{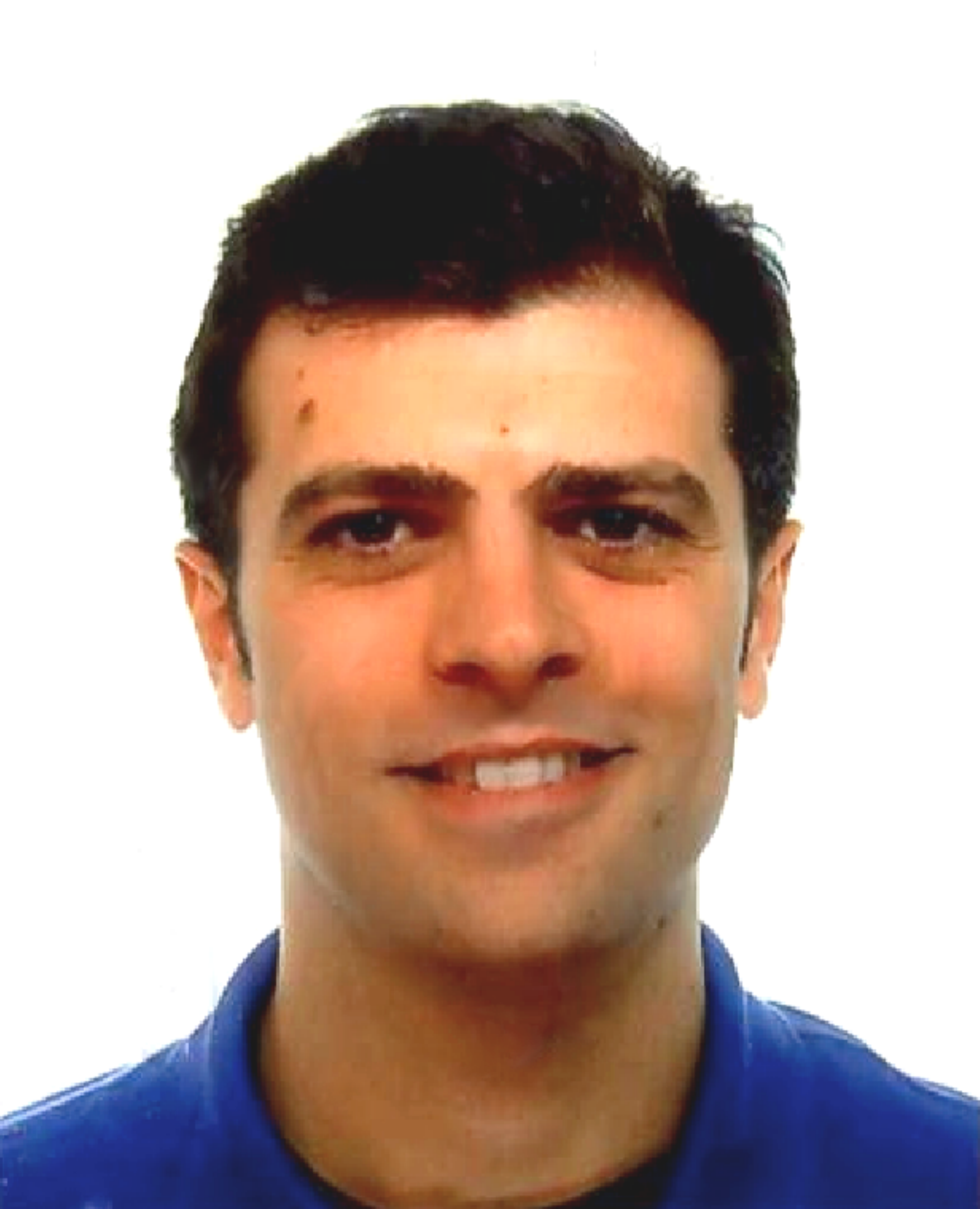}}
{\noindent \footnotesize{{\bf Battista Biggio}\
received the M.Sc. degree (Hons.) in Electronic Engineering and the Ph.D. degree in Electronic Engineering and Computer Science from the University of Cagliari, Italy, in 2006 and 2010. Since 2007, he has been with the Department of Electrical and Electronic Engineering, University of Cagliari, where he is currently an Assistant Professor. In 2011, he visited the University of Tuebingen, Germany, and worked on the security of machine learning to training data poisoning. His research interests include secure machine learning, multiple classifier systems, kernel methods, biometrics and computer security. Dr. Biggio serves as a reviewer for several international conferences and journals. He is a senior member of the IEEE and a member of the IAPR.}}
\vspace{2\baselineskip}}

\par\noindent 
\parbox[t]{\linewidth}{
\noindent\parpic{\includegraphics[height=1in]{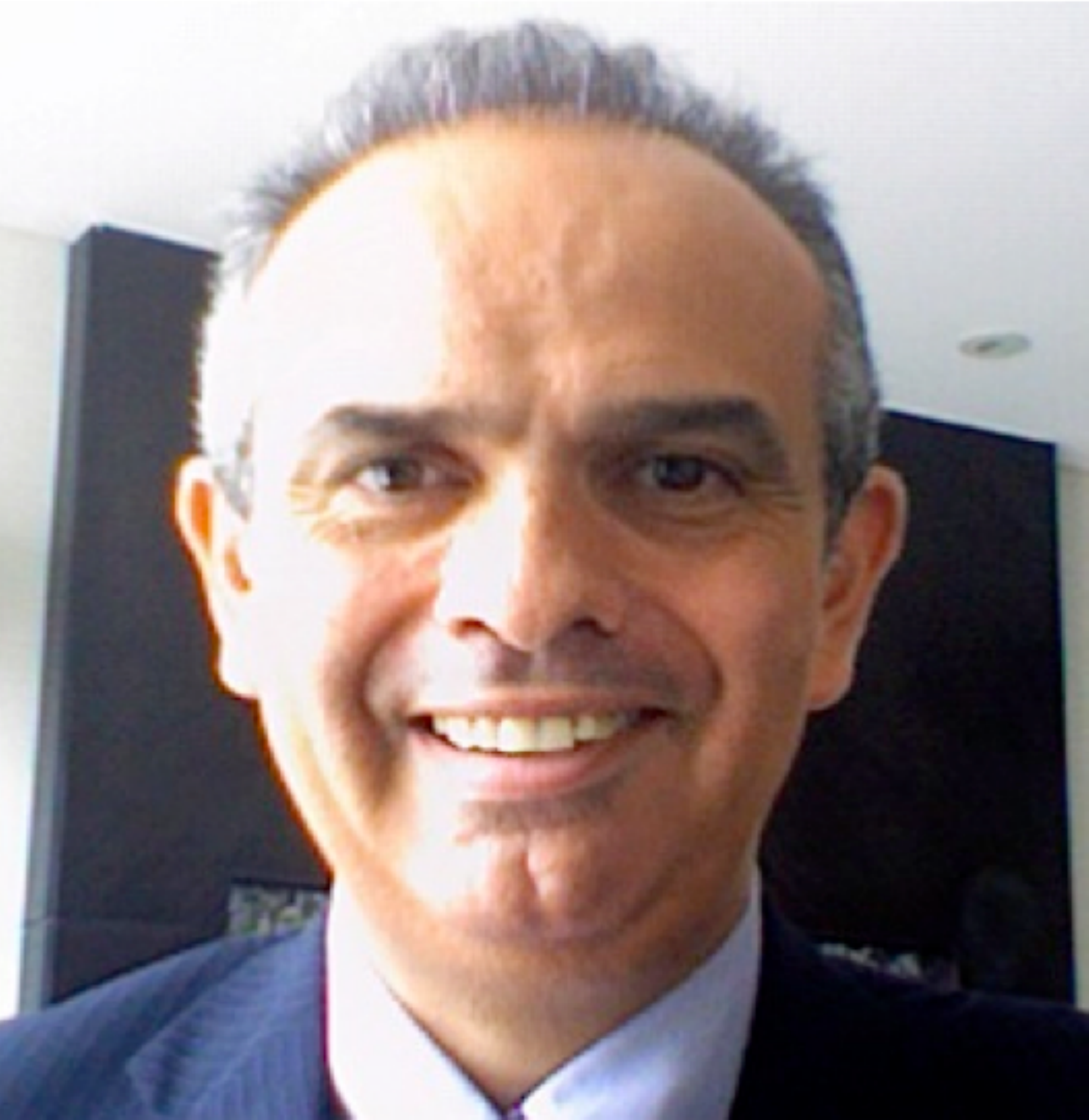}}
{\noindent \footnotesize{{\bf Fabio Roli}\
received his Ph.D. in Electronic Engineering from the University of Genoa, Italy. He was a research group member of the University of Genoa ('88-'94), and adjunct professor at the University of Trento ('93-'94). In 1995, he joined the Department of Electrical and Electronic Engineering of the University of Cagliari, where he is now professor of Computer Engineering and head of the research laboratory on pattern recognition and applications. His research activity is focused on the design of pattern recognition systems and their applications. He was a very active organizer of international conferences and workshops, and established the popular workshop series on multiple classifier systems. Dr. Roli is Fellow of the IEEE and of the IAPR.}}}

\balance
\end{document}